\documentclass[twoside,11pt]{article}

\usepackage{jmlr2e}

\usepackage{algorithm}
\usepackage{algorithmic}
\usepackage{overpic}
\usepackage{subfigure}

\usepackage{epsfig}
\usepackage{epstopdf}
\usepackage{graphicx}

\usepackage{amsmath,amssymb} 
\usepackage{multirow}
\usepackage{overpic}
\usepackage{color}

\usepackage{bbm}

\usepackage{enumitem}
\setlist[itemize]{leftmargin=0.8cm}
\setlist[enumerate]{leftmargin=0.8cm}

\usepackage{url}

\newcommand{\myendofproof}[0]{\hfill $\blacksquare$ \newline}

\firstpageno{1}

\input{def.set}

\begin{document}

\title{The Usual Suspects? \\ Reassessing Blame for VAE Posterior Collapse}

\author{\name Bin Dai \email daib13@mails.tsinghua.edu.cn \\
       \addr Institute for Advanced Study\\
       Tsinghua University\\
       Beijing, China
       \AND
       \name Ziyu Wang \email wzy196@gmail.com> \\
       \addr Department of Computer Science and Technology \\
       Tsinghua University\\
       Beijing, China
       \AND
      \name David Wipf \email davidwipf@gmail.com\\
      \addr Microsoft Research \\
      Beijing, China
}

\maketitle

\begin{abstract}
In narrow asymptotic settings Gaussian VAE models of continuous data have been shown to possess global optima aligned with ground-truth distributions.  Even so, it is well known that poor solutions whereby the latent posterior collapses to an uninformative prior are sometimes obtained in practice.  However, contrary to conventional wisdom that largely assigns blame for this phenomena on the undue influence of KL-divergence regularization, we will argue that posterior collapse is, at least in part, a direct consequence of bad local minima inherent to the loss surface of deep autoencoder networks.  In particular, we prove that even small nonlinear perturbations of affine VAE decoder models can produce such minima, and in deeper models, analogous minima can force the VAE to behave like an aggressive truncation operator, provably discarding information along all latent dimensions in certain circumstances.  Regardless, the underlying message here is not meant to undercut valuable existing explanations of posterior collapse, but rather, to refine the discussion and elucidate alternative risk factors that may have been previously underappreciated.
\end{abstract}

\section{Introduction}

The variational autoencoder (VAE) \citep{Kingma2014,Rezende2014} represents a powerful generative model of data points that are assumed to possess some complex yet unknown latent structure.  This assumption is instantiated via the marginalized distribution
\begin{equation} \label{eq:data_likelihood}
\textstyle{ p_{\theta}(\bx) = \int p_{\theta}(\bx|\bz) p(\bz) d\bz, }
\end{equation}
which forms the basis of prevailing VAE models.  Here $\bz \in \mathbb{R}^{\kappa}$ is a collection of unobservable latent factors of variation that, when drawn from the prior $p(\bz)$, are colloquially said to generate an observed data point $\bx \in \mathbb{R}^{d}$ through the conditional distribution $p_{\theta}(\bx|\bz)$.  The latter is controlled by parameters $\theta$ that can, at least conceptually speaking, be optimized by maximum likelihood over $p_{\theta}(\bx)$ given available training examples.


In particular, assuming $n$ training points $\bX = [\bx^{(1)},\ldots,\bx^{(n)}]$, maximum likelihood estimation is tantamount to minimizing the negative log-likelihood expression $\tfrac{1}{n}\sum_i -\log\left[ p_\theta\left(\bx^{(i)} \right) \right]$.  Proceeding further, because the marginalization over $\bz$ in (\ref{eq:data_likelihood}) is often intractable, the VAE instead minimizes a convenient variational upper bound given by ~ $\mathcal{L}(\theta,\phi)  \triangleq $
\vspace*{-0.3cm}
\begin{equation} \label{eq:VAE_objective}
 \tfrac{1}{n} \sum_{i=1}^n \left\{ - \mathbb{E}_{q_{\tiny \phi}\left(\bz|\bx^{(i)} \right)} \left[\log p_{\tiny \theta} \left(\bx^{(i)} | \bz  \right)  \right]  + \mathbb{KL}\left[q_\phi(\bz|\bx^{(i)}||p(\bz)\right] \right\} ~ \geq ~ \tfrac{1}{n}\sum_{i=1}^n -\log\left[ p_\theta\left(\bx^{(i)}\right) \right],
\end{equation}
with equality iff  $q_\phi(\bz|\bx^{(i)}) = p_\theta(\bz|\bx^{(i)})$ for all $i$.  The additional parameters $\phi$ govern the shape of the variational distribution $q_\phi(\bz|\bx)$ that is designed to approximate the true but often intractable latent posterior $p_\theta(\bz|\bx)$.

The VAE energy from (\ref{eq:VAE_objective}) is composed of two terms, a data-fitting loss that borrows the basic structure of an autoencoder (AE), and a KL-divergence-based regularization factor.  The former incentivizes assigning high probability to latent codes $\bz$ that facilitate accurate reconstructions of each $\bx^{(i)}$.  In fact, if $q_\phi(\bz|\bx)$ is a Dirac delta function, this term is exactly equivalent to a deterministic AE with data reconstruction loss defined by $-\log p_{\tiny \theta} \left(\bx | \bz  \right)$.  Overall, it is because of this association that $q_\phi(\bz|\bx)$ is generally referred to as the \emph{encoder} distribution, while $p_{\tiny \theta} \left(\bx | \bz  \right)$ denotes the \emph{decoder} distribution.  Additionally, the KL regularizer $\mathbb{KL}\left[q_\phi(\bz|\bx)||p(\bz)\right]$ pushes the encoder distribution towards the prior without violating the variational bound.

For continuous data, which will be our primary focus herein, it is typical to assume that
\begin{equation}
p(\bz) = \calN(\bz| {\bf 0}, \bI), ~~ p_{\tiny \theta}\left(\bx|\bz \right) = \calN(\bx | \bmu_x, \gamma \bI), ~ \mbox{and} ~ q_{\tiny \phi}\left(\bz|\bx \right) = \calN(\bz| \bmu_z, \bSigma_z ),
\end{equation}
where $\gamma > 0$ is a scalar variance parameter, while the Gaussian moments $\bmu_x \equiv \bmu_x\left(\bz; \theta  \right)$, $\bmu_z \equiv \bmu_z\left(\bx; \phi  \right)$, and $\bSigma_z \equiv \mbox{diag}[ \bsigma_z\left(\bx; \phi  \right)]^2$ are computed via feedforward neural network layers.  The encoder network parameterized by $\phi$  takes $\bx$ as an input and outputs $\bmu_z$ and $\bSigma_z$.  Similarly the decoder network parameterized by $\theta$ converts a latent code $\bz$ into $\bmu_x$.  Given these assumptions, the generic VAE objective from (\ref{eq:VAE_objective}) can be refined to
\begin{eqnarray} \label{eq:VAE_objective_general_Gaussian}
\calL(\theta,\phi) ~ = ~  \tfrac{1}{n} \sum_{i=1}^n \left\{ \mathbb{E}_{q_{\phi }\left(\bz|\bx^{(i)} \right)} \left[ \tfrac{1}{\gamma}\|\bx^{(i)} - \bmu_x \left( \bz; \theta \right)   \|_2^2  \right] \right. & &  \\
 && \hspace*{-6.4cm} \left.  + ~ d \log \gamma  +    \left\| \bsigma_z\left( \bx^{(i)}; \phi \right)  \right\|_2^2 - \log \left|\mbox{diag}\left[ \bsigma_z\left( \bx^{(i)}; \phi \right) \right]^2 \right| + \left\| \bmu_z\left( \bx^{(i)}; \phi \right)  \right\|_2^2   \right\}, \nonumber
\end{eqnarray}
excluding an inconsequential factor of $1/2$.  This expression can be optimized over using SGD and a simple reparameterization strategy \citep{Kingma2014,Rezende2014} to produce parameter estimates $\{\theta^*,\phi^*\}$.  Among other things, new samples approximating the training data can then be generated via the ancestral process $\bz^{new} \sim \calN(\bz| {\bf 0}, \bI)$ and $\bx^{new} \sim p_{\theta^*}(\bx | \bz^{new})$.


Although it has been argued that global minima of (\ref{eq:VAE_objective_general_Gaussian}) may correspond with the optimal recovery of ground truth distributions in certain asymptotic settings \citep{bin2019iclr}, it is well known that in practice, VAE models are at risk of converging to degenerate solutions where, for example, it may be that $q_{\phi}\left(\bz|\bx \right) = p(\bz)$.  This phenomena, commonly referred to as VAE posterior collapse \citep{he2019lagging,razavi2019preventing}, has been acknowledged and analyzed from a variety of different perspectives as we detail in Section \ref{sec:related_work}.  That being said, we would argue that there remains lingering ambiguity regarding the different types and respective causes of posterior collapse.  Consequently, Section \ref{sec:posterior_collapse} provides a useful taxonomy that will serve to contextualize our main technical contributions.  These include the following:
\begin{itemize}[leftmargin=*]
\item Building upon existing analysis of affine VAE decoder models, in Section \ref{sec:simple_cases} we prove that even arbitrarily small nonlinear activations can introduce suboptimal local minima exhibiting posterior collapse.
\item We demonstrate in Section \ref{sec:extrapolating} that if the encoder/decoder networks are incapable of sufficiently reducing the VAE reconstruction errors, even in a deterministic setting with no KL-divergence regularizer, there will exist an implicit lower bound on the optimal value of $\gamma$.  Moreover, we prove that if this $\gamma$ is sufficiently large, the VAE will behave like an aggressive thresholding operator, enforcing exact posterior collapse, i.e., $q_{\phi}\left(\bz|\bx \right) = p(\bz)$.
\item Based on these observations, we present experiments in Section \ref{sec:experiments} establishing that as network depth/capacity is increased, even for deterministic AE models with no regularization, reconstruction errors become worse.  This bounds the effective VAE trade-off parameter $\gamma$ such that posterior collapse is essentially inevitable.  Collectively then, we provide convincing evidence that posterior collapse is, at least in certain settings, the fault of deep AE local minima, and need not be exclusively a consequence of usual suspects such as the KL-divergence term.
\end{itemize}
We conclude in Section \ref{sec:conclusions} with practical take-home messages, and motivate the search for improved AE architectures and training regimes that might be leveraged by analogous VAE models.

\section{Recent Work and the Usual Suspects for Instigating Collapse} \label{sec:related_work}
Posterior collapse under various guises is one of the most frequently addressed topics related to VAE performance.  Depending on the context, arguably the most common and seemingly transparent suspect for causing collapse is the KL regularization factor that is obviously minimized by $q_{\phi}(\bz|\bx) = p(\bz)$.  This perception has inspired various countermeasures, including heuristic annealing of the KL penalty or KL warm-start \citep{bowman2015generating,huang2018improving,sonderby2016train}, tighter bounds on the log-likelihood \citep{burda2015importance,rezende2015variational}, more complex priors \citep{bauer2018resampled,tomczak2018vae}, modified decoder architectures \citep{cai2017multi,dieng2018avoiding,yeung2017tackling}, or efforts to explicitly disallow the prior from ever equaling the variational distribution \citep{razavi2019preventing}.  Thus far though, most published results do not indicate success generating high-resolution images, and in the majority of cases, evaluations are limited to small images and/or relatively shallow networks.  This suggests that there may be more nuance involved in pinpointing the causes and potential remedies of posterior collapse.  One notable exception though is the BIVA model from \citep{maaloe2019biva}, which employs a bidirectional  hierarchy of latent variables, in part to combat posterior collapse.  While improvements in NLL scores have been demonstrated with BIVA using relatively deep encoder/decoders, this model is significantly more complex and difficult to analyze.

On the analysis side, there have been various efforts to explicitly characterize posterior collapse in restricted settings.  For example, \citet{lucas2019understanding} demonstrate that if $\gamma$ is fixed to a sufficiently large value, then a VAE energy function with an affine decoder mean will have minima that overprune latent dimensions.  A related linearized approximation to the VAE objective is analyzed in \citep{rolinek2019variational}; however, collapsed latent dimensions are excluded and it remains somewhat unclear how the surrogate objective relates to the original.  Posterior collapse has also been associated with data-dependent decoder covariance networks $\bSigma_x(\bz;\theta) \neq \gamma \bI$ \citep{mattei2018leveraging}, which allows for degenerate solutions assigning infinite density to a single data point and a diffuse, collapsed density everywhere else.  Finally, from the perspective of training dynamics, \citep{he2019lagging} argue that a lagging inference network can also lead to posterior collapse.

\section{Taxonomy of Posterior Collapse} \label{sec:posterior_collapse}
Although there is now a vast literature on the various potential causes of posterior collapse, there remains ambiguity as to exactly what this phenomena is referring to.  In this regard, we believe that it is critical to differentiate five subtle yet quite distinct scenarios that could reasonably fall under the generic rubric of posterior collapse:
\begin{enumerate}[label=(\roman*),leftmargin=*]
\item Latent dimensions of $\bz$ that are not needed for providing good reconstructions of the training data are set to the prior, meaning $q_{\phi}(z_j|\bx) \approx p(z_j) = \calN(0,1)$ at any superfluous dimension $j$.  Along other dimensions $\bsigma_z^2$ will be near zero and $\bmu_z$ will provide a usable predictive signal leading to accurate reconstructions of the training data.  This case can actually be viewed as a desirable form of \emph{selective} posterior collapse that, as argued in \citep{bin2019iclr}, is a necessary (albeit not sufficient) condition for generating good samples.

\item The decoder variance $\gamma$ is not learned but fixed to a large value\footnote{Or equivalently, a KL scaling parameter such as used by the $\beta$-VAE \citep{higgins2017} is set too large.} such that the KL term from (\ref{eq:VAE_objective}) is overly dominant, forcing most or all dimensions of $\bz$ to follow the prior $\calN(0,1)$.  In this scenario, the actual global optimum of the VAE energy (conditioned on $\gamma$ being fixed) will lead to deleterious posterior collapse and the model reconstructions of the training data will be poor.  In fact, even the original marginal log-likelihood can potentially default to a trivial/useless solution if $\gamma$ is fixed too large, assigning a small marginal likelihood to the training data, provably so in the affine case \citep{lucas2019understanding}.

\item As mentioned previously, if the Gaussian decoder covariance is learned as a separate network structure (instead of simply $\bSigma_x(\bz; \theta) = \gamma \bI$), there can exist degenerate solutions that assign infinite density to a single data point and a diffuse, isotropic Gaussian elsewhere \citep{mattei2018leveraging}.  This implies that (\ref{eq:VAE_objective_general_Gaussian}) can be unbounded from below at what amounts to a posterior collapsed solution and bad reconstructions almost everywhere.

\item When powerful non-Gaussian decoders are used, and in particular those that can parameterize complex distributions regardless of the value of $\bz$ (e.g., PixelCNN-based \citep{van2016conditional}), it is possible for the VAE to assign high-probability to the training data even if $q_{\phi}(\bz|\bx) = p(\bz)$ \citep{alemi2017fixing,bowman2015generating,chen2016variational}.  This category of posterior collapse is quite distinct from categories (ii) and (iii) above in that, although the reconstructions are similarly poor, the associated NLL scores can still be good.

\item The previous four categories of posterior collapse can all be directly associated with emergent properties of the VAE \emph{global} minimum under various modeling conditions.  In contrast, a fifth type of collapse exists that is the explicit progeny of bad VAE \emph{local} minima.  More specifically, as we will argue shortly, when deeper encoder/decoder networks are used, the risk of converging to bad, overregularized solutions increases.

\end{enumerate}


The remainder of this paper will primarily focus on category (v), with brief mention of the other types for comparison purposes where appropriate.  Our rationale for this selection bias is that, unlike the others, category (i) collapse is actually advantageous and hence need not be mitigated.  In contrast, while category (ii) is undesirable, it be can be avoided by learning $\gamma$.  As for category (iii), this represents an unavoidable consequence of models with flexible decoder covariances capable of detecting outliers \citep{dai2017hidden}.  In fact, even simpler inlier/outlier decomposition models such as robust PCA are inevitably at risk for this phenomena \citep{Candes11}.  Regardless, when $\bSigma_z(\bx; \theta) = \gamma \bI$ this problem goes away.  And finally, we do not address category (iv) in depth simply because it is unrelated to the canonical Gaussian VAE models of continuous data that we have chosen to examine herein.  Regardless, it is still worthwhile to explicitly differentiate these five types and bare them in mind when considering attempts to both explain and improve VAE models.



%

\section{Insights from Simplified Cases} \label{sec:simple_cases}

Because different categories of posterior collapse can be impacted by different global/local minima structures, a useful starting point is a restricted setting whereby we can comprehensively characterize all such minima.  For this purpose, we first consider a VAE model with the decoder network set to an affine function.  As is often assumed in practice, we choose $\bSigma_x = \gamma \bI$, where $\gamma > 0$ is a scalar parameter within the parameter set $\theta$. In contrast, for the mean function we choose $\bmu_x =  \bW_x \bz + \bb_x$ for some weight matrix $\bW_x$ and bias vector $\bb_x$.  The encoder can be arbitrarily complex (although the optimal structure can be shown to be affine as well).



Given these simplifications, and assuming the training data has $r \geq \kappa$ nonzero singular values, it has been demonstrated that at any global optima, the columns of $\bW_x$ will correspond with the first $\kappa$ principal components of $\bX$ provided that we simultaneously learn $\gamma$ or set it to the optimal value (which is available in closed form) \citep{dai2017hidden,lucas2019understanding,Tipping1999}.  Additionally, it has also be shown that no spurious, suboptimal local minima will exist.  Note also that if $r < \kappa$ the same basic conclusions still apply; however, $\bW_x$ will only have $r$ nonzero columns, each corresponding with a different principal component of the data.  The unused latent dimensions will satisfy $q_{\phi}(\bz|\bx) = \calN({\bf 0}, \bI)$, which represents the canonical form of the benign category (i) posterior collapse.  Collectively, these results imply that if we converge to any local minima of the VAE energy, we will obtain the best possible linear approximation to the data using a minimal number of latent dimensions, and malignant posterior collapse is not an issue, i.e., categories (ii)-(v) will not arise.

Even so, if instead of learning $\gamma$, we choose a fixed value that is larger than any of the significant singular values of $\bX \bX^{\top}$, then category (ii) posterior collapse can be inadvertently introduced.  More specifically, let $\tilde{r}_{\gamma}$ denote the number of such singular values that are smaller than some fixed $\gamma$ value.  Then along $\kappa-\tilde{r}_{\gamma}$ latent dimensions $q_{\phi}(\bz|\bx) = \calN({\bf 0}, \bI)$, and the corresponding columns of $\bW_x$ will be set to zero at the global optima (conditioned on this fixed $\gamma$), regardless of whether or not these dimensions are necessary for accurately reconstructing the data.  And it has been argued that the risk of this type of posterior collapse at a conditionally-optimal global minimum will likely be inherited by deeper models as well \citep{lucas2019understanding}, although learning $\gamma$ can ameliorate this problem.

Of course when we move to more complex architectures, the risk of bad \emph{local} minima or other suboptimal stationary points becomes a new potential concern, and it is not clear that the affine case described above contributes to reliable, predictive intuitions.  To illustrate this point, we will now demonstrate that the introduction of an arbitrarily small nonlinearity can nonetheless produce a pernicious local minimum  that exhibits category (v) posterior collapse.  For this purpose, we assume the decoder mean function
\begin{equation}  \label{eq:nonlinear_decoder_mean}
\bmu_x =  \pi_{\alpha}\left( \bW_x \bz \right) + \bb_x, ~ \mbox{with} ~ \pi_{\alpha}(u) \triangleq \mbox{sign}(u) \left(\left| u \right| - \alpha   \right)_+, ~ \alpha \geq 0.
\end{equation}
The function $\pi_{\alpha}$ is nothing more than a soft-threshold operator as is commonly used in neural network architectures designed to reflect unfolded iterative algorithms for representation learning \citep{Gregor10,Sprechmann15}.  In the present context though, we choose this nonlinearity largely because it allows (\ref{eq:nonlinear_decoder_mean}) to reflect arbitrarily small perturbations away from a strictly affine model, and indeed if $\alpha = 0$ the exact affine model is recovered.  Collectively, these specifications lead to the parameterization $\theta = \{\bW_x, \bb_x, \gamma \}$ and $\phi = \{\bmu_z^{(i)}, \bsigma_z^{(i)} \}_{i=1}^n$ and energy (excluding irrelevant scale factors and constants) given by 
\begin{eqnarray} \label{eq:VAE_AffinePlus_Cost}
\calL(\theta,\phi) ~ = ~ \sum_{i=1}^n \left\{ \mathbb{E}_{q_{\tiny \phi}\left(\bz|\bx^{(i)} \right)} \left[ \tfrac{1}{\gamma} \left\| \bx^{(i)} - \pi_{\alpha}\left( \bW_x \bz \right)   - \bb_x  \right\|_2^2  \right] \right. & &  \\
 && \hspace*{-4.0cm} \left.  + d \log \gamma  +  \left\| \bsigma_z^{(i)} \right\|_2^2  - \log \left| \mbox{diag}\left[ \bsigma_z^{(i)} \right]^2 \right|  + \left\| \bmu_z^{(i)}  \right\|_2^2  \right\}, \nonumber
\end{eqnarray}
where $\bmu_z^{(i)}$ and $\bsigma_z^{(i)}$ denote arbitrary encoder moments for data point $i$ (this is consistent with the assumption of an arbitrarily complex encoder as used in previous analysis of affine decoder models).  Now define $\bar{\gamma} \triangleq \tfrac{1}{n d} \sum_i \| \bx^{(i)} - \bar{\bx} \|_2^2$, with $\bar{\bx} \triangleq \tfrac{1}{n} \sum_i  \bx^{(i)}$.  We then have the following result (all proofs are deferred to the appendices):
\begin{proposition} \label{prop:non_affine_special_case}
For any $\alpha > 0$, there will always exist data sets $\bX$ such that (\ref{eq:VAE_AffinePlus_Cost}) has a global minimum that perfectly reconstructs the training data, but also a bad local minimum characterized by
\begin{equation} \label{eq:non_affine_bad_local_minimum}
q_{\phi}(\bz | \bx) = \calN(\bz | {\bf 0}, \bI) ~~~ \mbox{and} ~~~ p_{\theta}(\bx) = \calN(\bx | \bar{\bx},\bar{\gamma} \bI ).
\end{equation}
\end{proposition}
Hence the moment we allow for nonlinear (or more precisely, non-affine) decoders there can exist a poor local minimum, across all parameters including a learnable $\gamma$, that exhibits category (v) posterior collapse.\footnote{This result mirrors related efforts examining linear DNNs, where it has been previously demonstrated that under certain conditions, all local minima are globally optimal \citep{Kawaguchi2016}, while small nonlinearities can induce bad local optima \citep{yun2018small}.  However, the loss surface of these models is completely different from a VAE, and hence we view Proposition \ref{prop:non_affine_special_case} as a complementary result.}  In other words, no predictive information about $\bx$ passes through the latent space, and a useless/non-informative distribution $p_{\theta}(\bx)$ emerges that is incapable of assigning high probability to the data (except obviously in the trivial degenerate case where all the data points are equal to the empirical mean $\bar{\bx}$).  We will next investigate the degree to which such concerns can influence behavior in arbitrarily deep architectures.



\section{Extrapolating to Practical Deep Architectures} \label{sec:extrapolating}

Previously we have demonstrated the possibility of local minima aligned with category (v) posterior collapse the moment we allow for decoders that deviate ever so slightly from an affine model.  But nuanced counterexamples designed for proving technical results notwithstanding, it is reasonable to examine what realistic factors are largely responsible for leading optimization trajectories towards such potential bad local solutions.  For example, is it merely the strength of the KL regularization term, and if so, why can we not just use KL warm-start to navigate around such points?  In this section we will elucidate a deceptively simple, alternative risk factor that will be corroborated empirically in Section \ref{sec:experiments}.

From the outset, we should mention that with deep encoder/decoder architectures commonly used in practice, a stationary point can more-or-less always exist at solutions exhibiting posterior collapse.  As a representative and ubiquitous example, please see Appendix \ref{sec:stationary_point_example}.  But of course without further details, this type of stationary point could conceivably manifest as a saddle point (stable or unstable), a local maximum, or a local minimum.  For the strictly affine decoder model mentioned in Section \ref{sec:simple_cases}, there will only be a harmless unstable saddle point at any collapsed solution (the Hessian has negative eigenvalues).  In contrast, for the special nonlinear case elucidated via Proposition \ref{prop:non_affine_special_case} we can instead have a bad local minima.  We will now argue that as the depth of common feedforward architectures increases, the risk of converging to category (v)-like solutions with most or all latent dimensions stuck at bad stationary points can also increase.

Somewhat orthogonal to existing explanations of posterior collapse, our basis for this argument is not directly related to the VAE KL-divergence term.  Instead, we consider a deceptively simple yet potentially influential alternative:  \emph{Unregularized, deterministic AE models can have bad local solutions with high reconstruction errors when sufficiently deep.  This in turn can directly translate to category (v) posterior collapse when training a corresponding VAE model with a matching deep architecture.}  Moreover, to the extent that this is true, KL warm-start or related countermeasures will likely be ineffective in avoiding such suboptimal minima.  We will next examine these claims in greater depth followed by a discussion of practical implications.

\subsection{From Deeper Architectures to Inevitable Posterior Collapse}


%
%
%


Consider the deterministic AE model formed by composing the encoder mean $\bmu_x \equiv \bmu_x\left( \cdot ; \theta   \right)$ and decoder mean $\bmu_z \equiv \bmu_z\left( \cdot ; \phi   \right)$ networks from a VAE model, i.e., reconstructions $\hat{\bx}$ are computed via $\hat{\bx} =  \bmu_x\left[ \bmu_z\left( \bx; \phi \right); \theta \right]$.  We then train this AE to minimize the squared-error loss $\tfrac{1}{n d} \sum_{i=1}^n \left\|\bx^{(i)} - \hat{\bx}^{(i)}  \right\|_2^2$, producing parameters $\{\theta_{ae}, \phi_{ae}\}$.  Analogously, the corresponding VAE trained to minimize (\ref{eq:VAE_objective_general_Gaussian}) arrives at a parameter set denoted $\{\theta_{vae}, \phi_{vae}\}$.  In this scenario, it will typically follow that
\begin{equation} \label{eq:reconstruction_error_bound}
\tfrac{1}{n d} \sum_{i=1}^n \left\|\bx^{(i)} - \bmu_x\left[ \bmu_z\left( \bx^{(i)}; \phi_{ae} \right); \theta_{ae} \right] \right\|_2^2 \leq \tfrac{1}{nd} \sum_{i=1}^n \mathbb{E}_{q_{\phi_{vae}}\left(\bz|\bx^{(i)} \right)} \left[ \|\bx^{(i)} - \bmu_x \left( \bz; \theta_{vae} \right)   \|_2^2  \right],
\end{equation}
meaning that the deterministic AE reconstruction error will generally be smaller than the stochastic VAE version.  Note that if $\bsigma_z^2 \rightarrow 0$, the VAE defaults to the same deterministic encoder as the AE and hence will have identical representational capacity; however, the KL regularization prevents this from happening, and any $\bsigma_z^2 > 0$ can only make the reconstructions worse.\footnote{Except potentially in certain contrived adversarial conditions that do not represent practical regimes.}  Likewise, the KL penalty factor $\| \bmu_z^2 \|_2^2$ can further restrict the effective capacity and increase the reconstruction error of the training data.  Beyond these intuitive arguments, we have never empirically found a case where (\ref{eq:reconstruction_error_bound}) does not hold (see Section \ref{sec:experiments} for examples).


We next define the set
\begin{equation}
\calS_{\varepsilon} ~ \triangleq ~ \left\{~\theta,\phi ~:~ \tfrac{1}{n d} \sum_{i=1}^n \left\|\bx^{(i)} - \hat{\bx}^{(i)}  \right\|_2^2  \leq \varepsilon   \right\}
\end{equation}
for any $\epsilon > 0$.  Now suppose that the chosen encoder/decoder architecture is such that with high probability, achievable optimization trajectories (e.g., via SGD or related) lead to parameters $\{\theta_{ae}, \phi_{ae}\} \notin \calS_{\varepsilon}$, i.e., $\mbox{Prob}\left( \{\theta_{ae}, \phi_{ae}\} \in \calS_{\varepsilon} \right) \approx 0$.  It then follows that the optimal VAE noise variance denoted $\gamma^*$, when conditioned on practically-achievable values for other network parameters, will satisfy
\begin{equation} \label{eq:gamma_lower_bound}
\gamma^* ~ =  ~ \tfrac{1}{nd} \sum_{i=1}^n \mathbb{E}_{q_{\phi_{vae}}\left(\bz|\bx^{(i)} \right)} \left[ \|\bx^{(i)} - \bmu_x \left( \bz; \theta_{vae} \right)   \|_2^2  \right] ~ \geq ~ \varepsilon.
\end{equation}
The equality in (\ref{eq:gamma_lower_bound}) can be confirmed by simply differentiating the VAE cost w.r.t. $\gamma$ and equating to zero, while the inequality comes from (\ref{eq:reconstruction_error_bound}) and the fact that $\{\theta_{ae}, \phi_{ae}\} \notin \calS_{\varepsilon}$.


From inspection of the VAE energy from (\ref{eq:VAE_objective_general_Gaussian}), it is readily apparent that larger values of $\gamma$ will discount the data-fitting term and therefore place greater emphasis on the KL divergence.  Since the latter is minimized when the latent posterior equals the prior, we might expect that whenever $\varepsilon$ and therefore $\gamma^*$ is increased per (\ref{eq:gamma_lower_bound}), we are at a greater risk of nearing collapsed solutions.  But the nature of this approach is not at all transparent, and yet this subtlety has important implications for understanding the VAE loss surface in regions at risk of posterior collapse.

For example, one plausible hypothesis is that only as $\gamma^* \rightarrow \infty$ do we risk full category (v) collapse.  If this were the case, we might have less cause for alarm since the reconstruction error and by association $\gamma^*$ will typically be bounded from above at any local minimizer.  However, we will now demonstrate that even finite values can exactly collapse the posterior.  In formally showing this, it is helpful to introduce a slightly narrower but nonetheless representative class of VAE models.

Specifically, let $f\left(\bmu_z, \bsigma_z, \theta, \bx^{(i)} \right) \triangleq \mathbb{E}_{q_{\phi}\left(\bz|\bx^{(i)} \right)} \left[ \|\bx^{(i)} - \bmu_x \left( \bz; \theta \right)  \|_2^2  \right]$, i.e., the VAE data term evaluated at a single data point without the $1/\gamma$ scale factor.  We then define a \emph{well-behaved VAE} as a model with energy function (\ref{eq:VAE_objective_general_Gaussian}) designed such that $\nabla_{\mu_z} f\left(\bmu_z, \bsigma_z, \theta, \bx^{(i)} \right)$ and $\nabla_{\sigma_z} f\left(\bmu_z, \bsigma_z, \theta, \bx^{(i)} \right)$ are Lipschitz continuous gradients for all $i$.  Furthermore, we specify a \emph{non-degenerate decoder} as any $\bmu_x ( \bz; \theta = \tilde{\theta} )$ with $\theta$ set to a $\tilde{\theta}$ value such that $\nabla_{\sigma_z} f\left(\bmu_z, \bsigma_z, \tilde{\theta}, \bx^{(i)}\right) \geq c$ for some constant $c > 0$ that can be arbitrarily small.  This ensures that $f$ is an increasing function of $\bsigma_z$, a quite natural stipulation given that increasing the encoder variance will generally only serve to corrupt the reconstruction, unless of course the decoder is completely blocking the signal from the encoder.  In the latter degenerate situation, it would follow that  $\nabla_{\mu_z} f\left(\bmu_z, \bsigma_z, \theta, \bx^{(i)} \right) = \nabla_{\sigma_z} f\left(\bmu_z, \bsigma_z, \theta, \bx^{(i)} \right) = 0$, which is more-or-less tantamount to category (v) posterior collapse.

%

Based on these definitions, we can now present the following:
\begin{proposition} \label{prop:guaranteed_posterior_collapse}
For any well-behaved VAE with arbitrary, non-degenerate decoder $\bmu_x(\bz;\theta = \tilde{\theta})$, there will always exist a $\gamma' < \infty$ such that the trivial solution $\bmu_x(\bz;\theta \neq \tilde{\theta}) = \bar{\bx}$ and $q_{\phi}(\bz|\bx) = p(\bz)$ will have lower cost.
\end{proposition}

Around any evaluation point, the sufficient condition we applied to demonstrate posterior collapse (see proof details) can also be achieved with some $\gamma'' < \gamma'$ if we allow for partial collapse, i.e.,  $q_{\phi^*}(z_j | \bx) = p(z_j)$ along some but not all latent dimensions $j \in \{1,\ldots,\kappa \}$.  Overall, the analysis loosely suggests that the number of dimensions vulnerable to exact collapse will increase monotonically with $\gamma$.

Proposition \ref{prop:guaranteed_posterior_collapse} also provides evidence that the VAE behaves like a strict thresholding operator, completely shutting off latent dimensions using a finite value for $\gamma$.  This is analogous to the distinction between using the $\ell_1$ versus $\ell_2$ norm for solving regularized regression problems of the standard form ~$\min_{\bu} \| \bx - \bA \bu \|_2^2 + \gamma \hspace*{0.05cm} \eta(\bu)$, where $\bA$ is a design matrix and $\eta$ is a penalty function.  When $\eta$ is the $\ell_1$ norm, some or all elements of $\bu$ can be pruned to exactly zero with a sufficiently large but finite $\gamma$ \cite{zhao2006model}.  In contrast, when the $\ell_2$ norm is applied, the coefficients will be shrunk to smaller values but never pushed all the way to zero unless $\gamma \rightarrow \infty$.

\subsection{Practical Implications} \label{sec:practical_implications}

In aggregate then, if the AE base model displays unavoidably high reconstruction errors, this implicitly constrains the corresponding VAE model to have a large  optimal $\gamma$ value, which can potentially lead to undesirable posterior collapse per Proposition \ref{prop:guaranteed_posterior_collapse}.  In Section \ref{sec:experiments} we will demonstrate empirically that training unregularized AE models can become increasingly difficult and prone to bad local minima (or at least bad stable stationary points) as the depth increases; and this difficulty can persist even with counter-measures such as skip connections.  Therefore, from this vantage point we would argue that it is \emph{ the AE base architecture that is effectively the guilty party when it comes to category (v) posterior collapse}.

The perspective described above also helps to explain why heuristics like KL warm-start are not always useful for improving VAE performance.  With the standard Gaussian model (\ref{eq:VAE_objective_general_Gaussian}) considered herein, KL warm-start amounts to adopting a pre-defined schedule for incrementally increasing $\gamma$ starting from a small initial value, the motivation being that a small $\gamma$ will steer optimization trajectories away from overregularized solutions and posterior collapse.

However, regardless of how arbitrarily small $\gamma$ may be fixed at any point during this process, the VAE reconstructions are not likely to be better than the analogous deterministic AE (which is roughly equivalent to forcing $\gamma = 0$ within the present context).  This implies that there can exist an \emph{implicit} $\gamma^*$ as computed by (\ref{eq:gamma_lower_bound}) that can be significantly larger such that, even if KL warm-start is used, the optimization trajectory may well lead to a collapsed posterior stationary point that has this $\gamma^*$ as the optimal value in terms of minimizing the VAE cost with other parameters fixed.  Note that if full posterior collapse does occur, the gradient from the KL term will equal zero and hence, to be at a stationary point it must be that the data term gradient is also zero.  In such situations, varying $\gamma$ manually will not impact the gradient balance anyway.

%
%



\section{Empirical Assessments} \label{sec:experiments}

\begin{figure}[t!]
    \centering
    \includegraphics[width=0.32\textwidth]{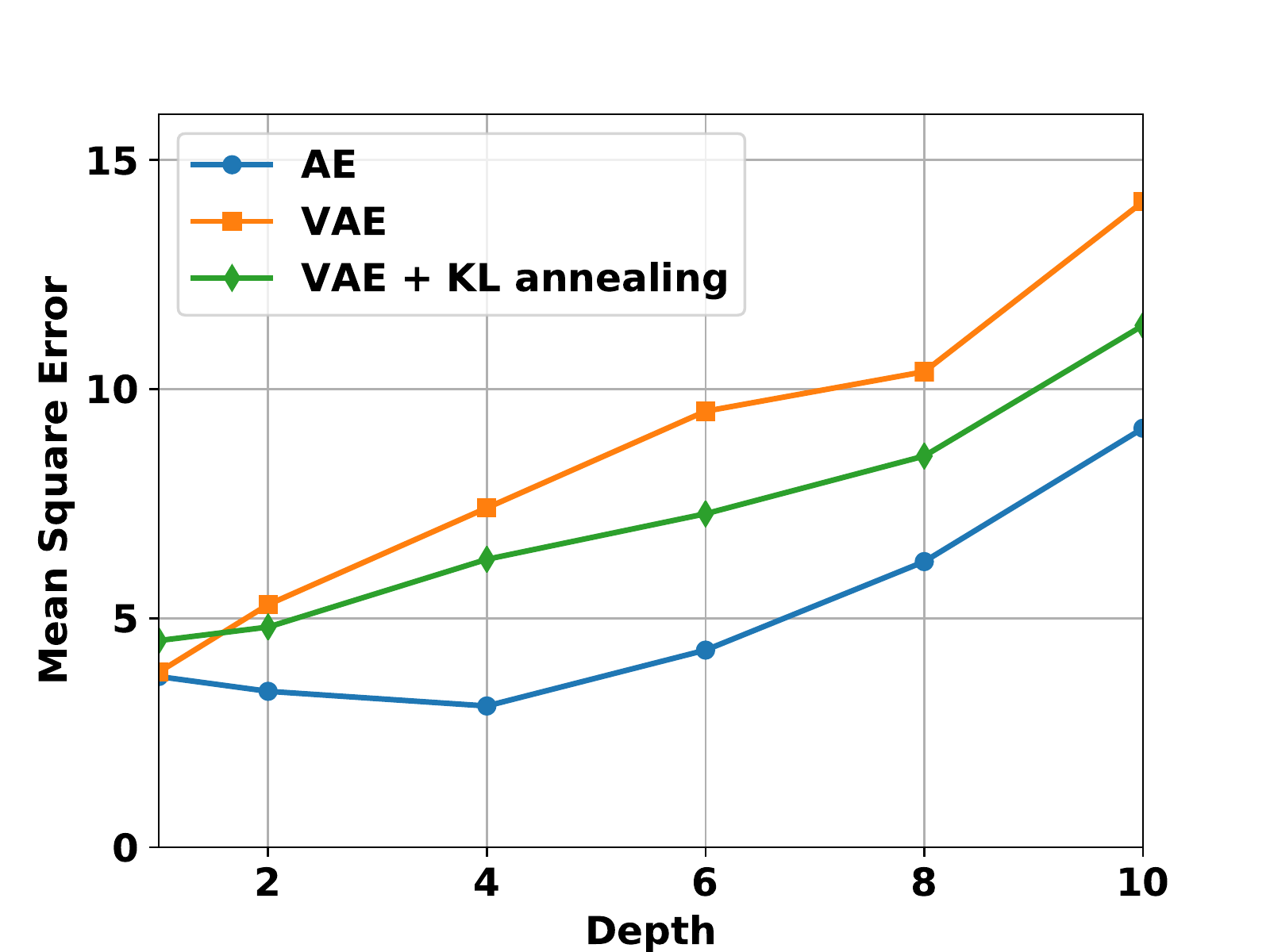}
    \includegraphics[width=0.32\textwidth]{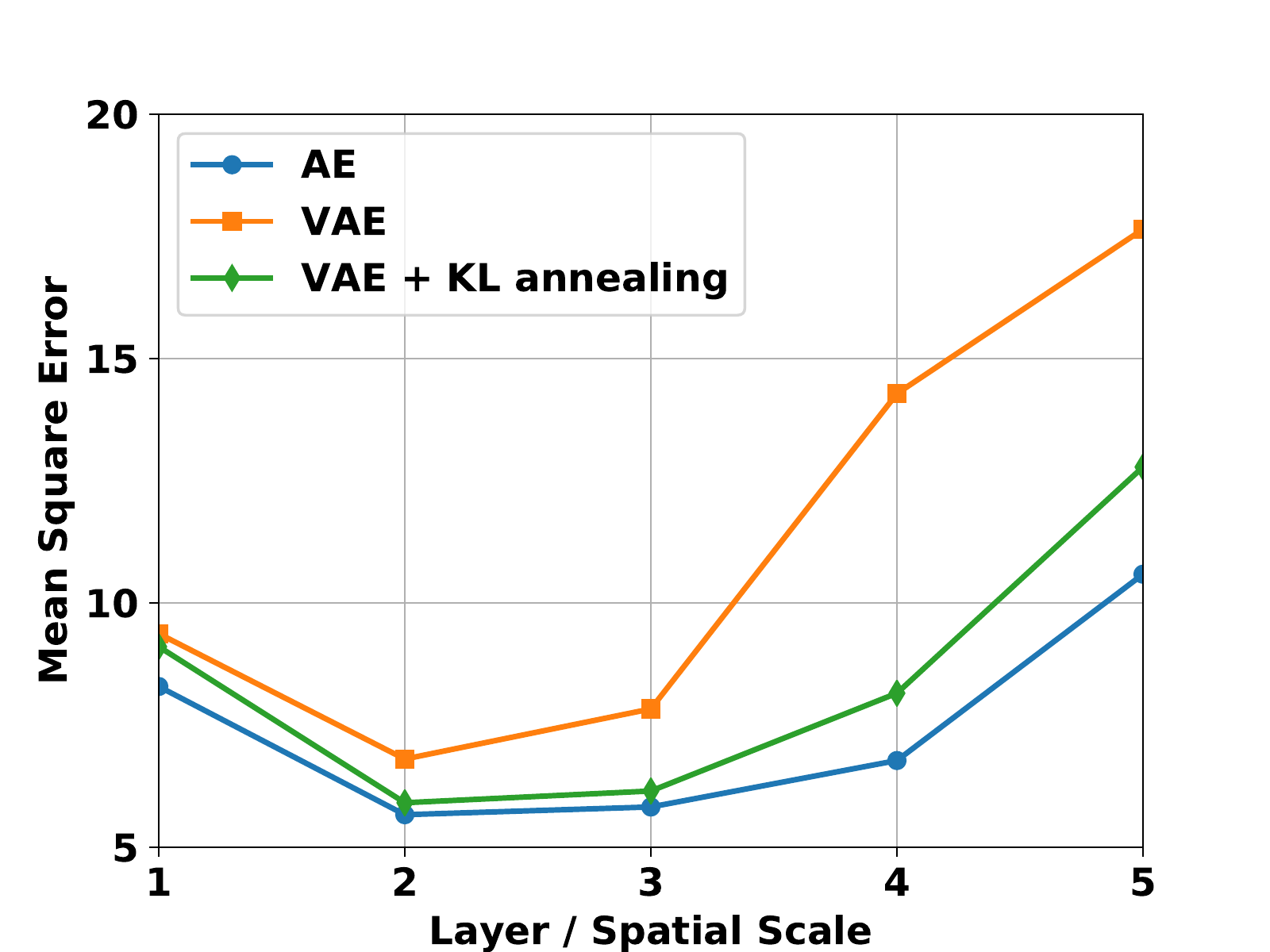}
    \includegraphics[width=0.32\textwidth]{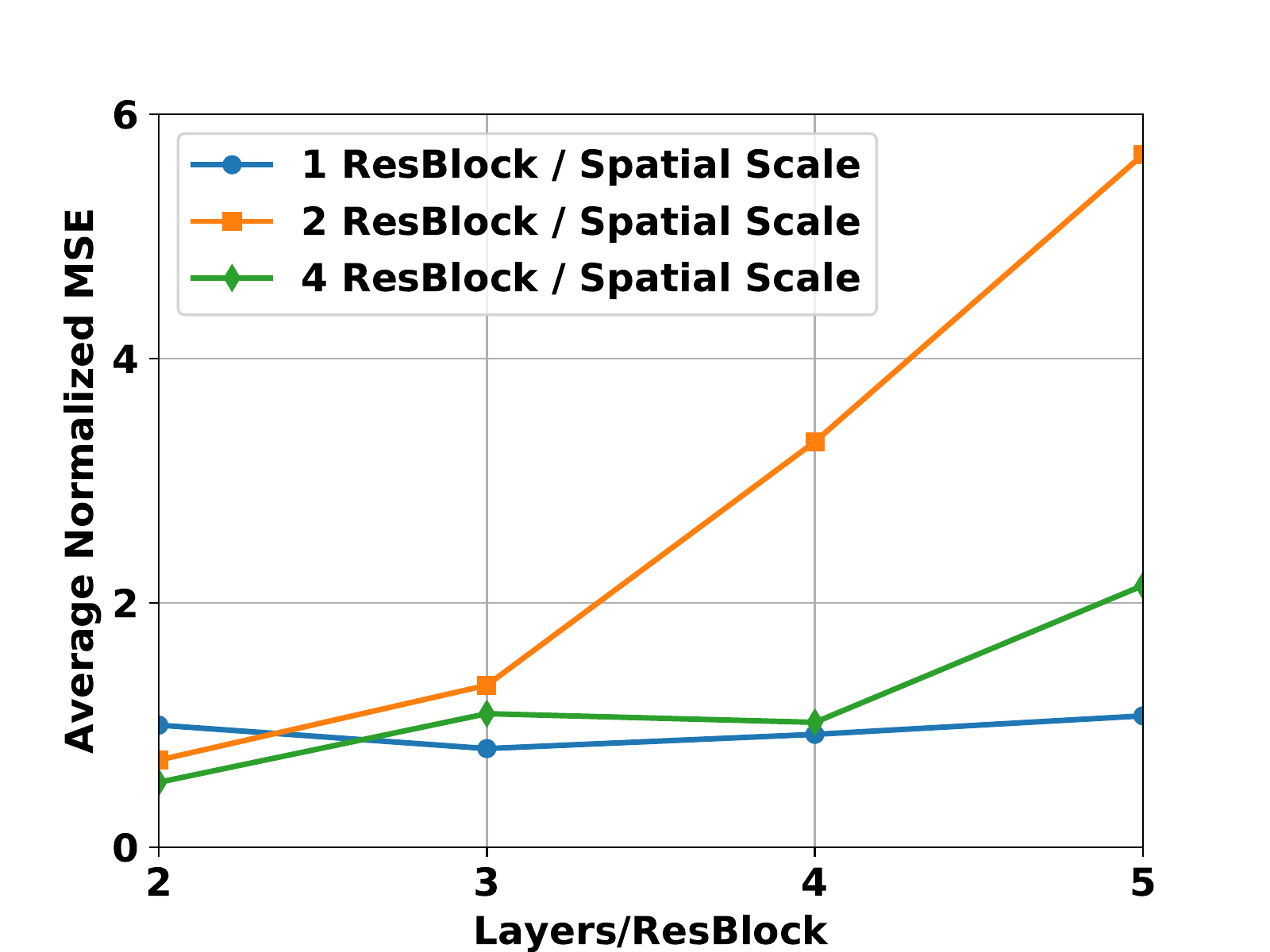}
    \caption{Reconstruction errors for various encoder/decoder models of varying complexity. \emph{Left}: Fully connected networks with different depths trained on Fashion-MNIST. \emph{Middle}: Convolution networks with increasing depth/\# of spatial scales trained on Cifar100.  \emph{Right}: Averaged AE results from residual networks with varying number of residual blocks and block depth trained on SVHN, Cifar10, Cifar100 and CelebA.  In all plots, once the encoder/decoder complexity is sufficiently high, the reconstruction errors begin to increase.}
    \label{fig:recon_err}
\end{figure}

In this section we empirically demonstrate the existence of bad AE local minima with high reconstruction errors at increasing depth, as well as the association between these bad minima and imminent VAE posterior collapse.  For this purpose, we first train fully connected AE and VAE models with $1$, $2$, $4$, $6$, $8$ and $10$ hidden layers on the Fashion-MNIST dataset~\citep{xiao2017fashion}. Each hidden layer is $512$-dimensional and followed by ReLU activations (see Appendix \ref{sec:experimental_details} for further details). The reconstruction error is shown in Figure~\ref{fig:recon_err}(\emph{left}). As the depth of the network increases, the reconstruction error of the AE model first decreases because of the increased capacity. However, when the network becomes too deep, the error starts to increase, indicating convergence to a bad local minima (or at least stable stationary point/plateau) that is unrelated to KL-divergence regularization. The reconstruction error of a VAE model is always worse than that of the corresponding AE model as expected.  Moreover, while KL warm-start/annealing can help to improve the VAE reconstructions to some extent, performance is still worse than the AE as expected.

We next train AE and VAE models using a more complex convolutional network on Cifar100 data~\citep{krizhevsky2009learning}. At each spatial scale, we use $1$ to $5$ convolution layers followed by ReLU activations. We also apply $2\times2$ max pooling to downsample the feature maps to a smaller spatial scale in the encoder and use a transposed convolution layer to upscale the feature map in the decoder. The reconstruction errors are shown in Figure~\ref{fig:recon_err}(\emph{middle}). Again, the trend is similar to the fully-connected network results.  See Appendix \ref{sec:experimental_details} for an additional ImageNet example.

It has been argued in the past that skip connections can increase the mutual information between observations $\bx^{(i)}$ and the inferred latent variables $\bz$ \citep{dieng2018avoiding}, reducing the risk of posterior collapse.  And it is well-known that ResNet architectures based on skip connections can improve performance on numerous recognition tasks~\citep{he2015deep}.  To this end, we train a number of AE models using ResNet-inspired encoder/decoder architectures on multiple datasets including Cifar10, Cifar100, SVHN and CelebA. Similar to the convolution network structure from above, we use $1$, $2$, and $4$ residual blocks within each spatial scale. Inside each block, we apply $2$ to $5$ convolution layers. For aggregate comparison purposes, we normalize the reconstruction error obtained on each dataset by dividing it with the corresponding error produced by the most shallow network structure ($1$ residual block with $2$ convolution layers). We then average the normalized reconstruction errors over all four datasets. The average normalized errors are shown in Figure~\ref{fig:recon_err}(\emph{right}), where we observe that adding more convolution layers inside each residual block can increase the reconstruction error when the network is too deep. Moreover, adding more residual blocks can also lead to higher reconstruction errors.  And empirical results obtained using different datasets and networks architectures, beyond the conditions of Figure~\ref{fig:recon_err}, also show a general trend of increased reconstruction error once the effective depth is sufficiently deep.  

We emphasize that in all these models, as the network complexity/depth increases, the simpler models are always contained within the capacity of the larger ones.  Therefore, because the reconstruction error on the training data is becoming worse, it must be the case that the AE is becoming stuck at bad local minima or plateaus.  Again since the AE reconstruction error serves as a probable lower bound for that of the VAE model, a deeper VAE model will likely suffer the same problem, only exacerbated by the KL-divergence term in the form of posterior collapse.   This implies that there will be more $\bsigma_z$ values moving closer to $1$ as the VAE model becomes deeper; similarly $\bmu_z$ values will push towards 0. The corresponding dimensions will encode no information and become completely useless.

\begin{figure}[t!]
    \centering
    \includegraphics[width=0.32\textwidth]{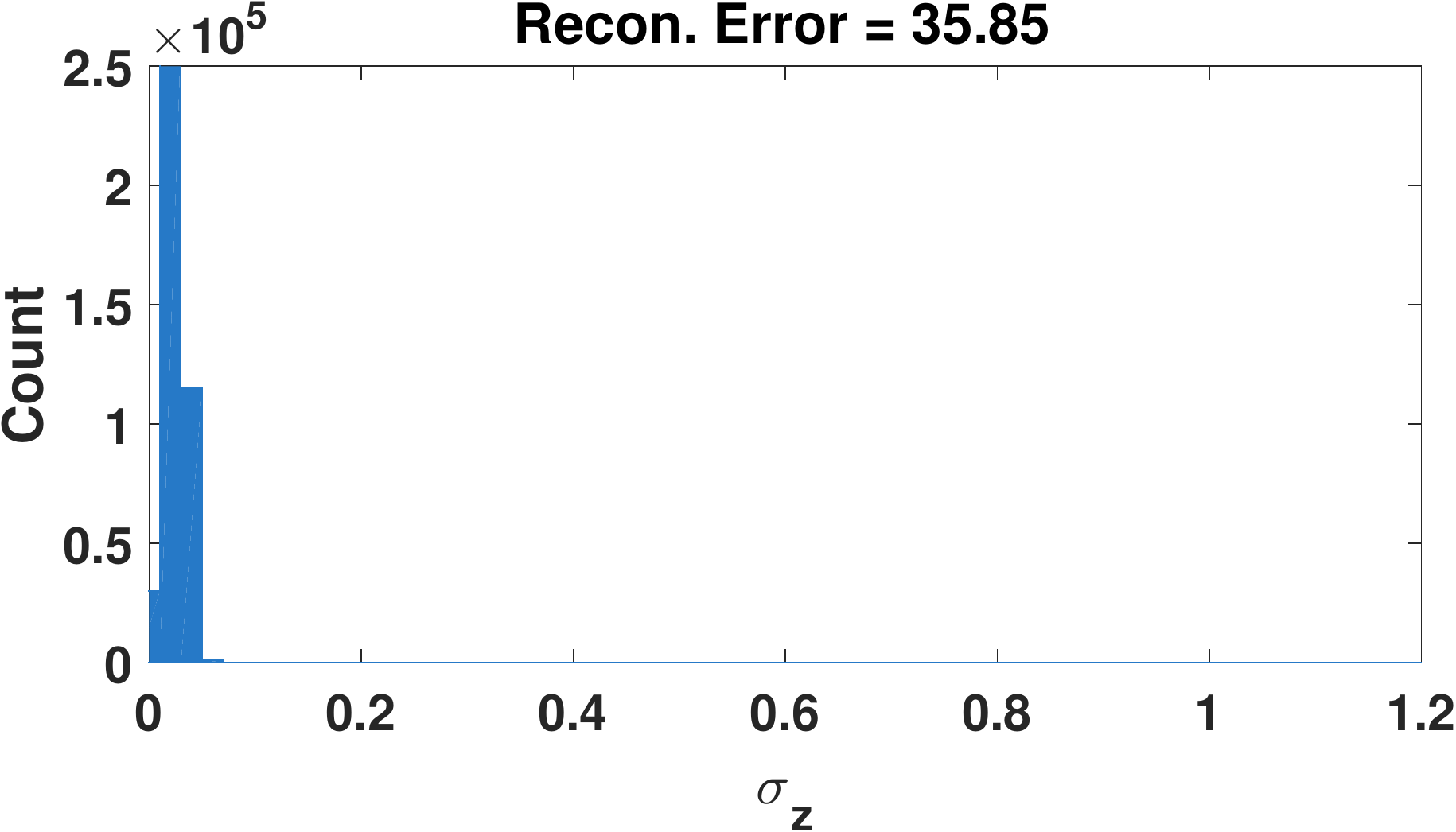}
    \includegraphics[width=0.32\textwidth]{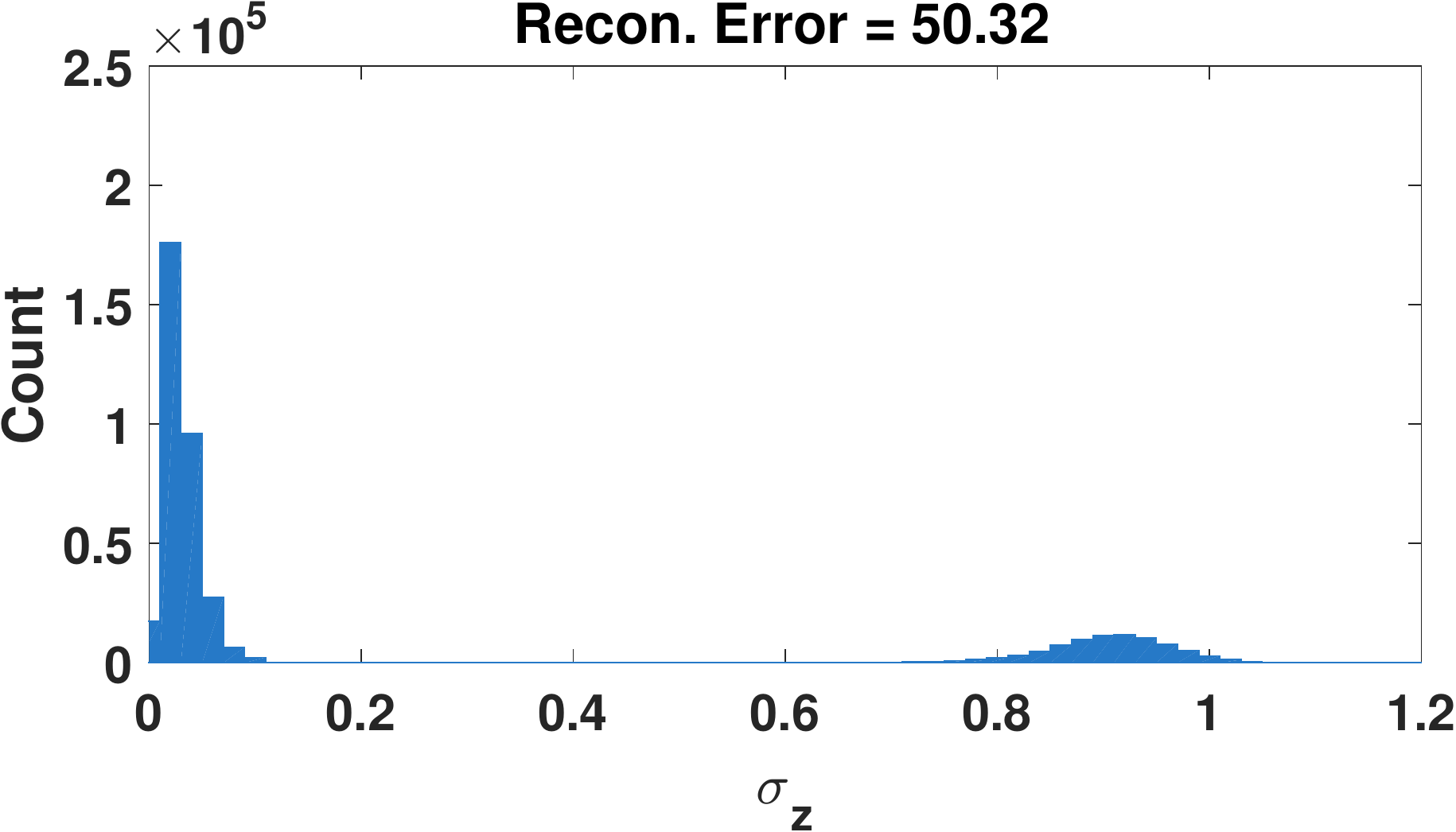}
    \includegraphics[width=0.32\textwidth]{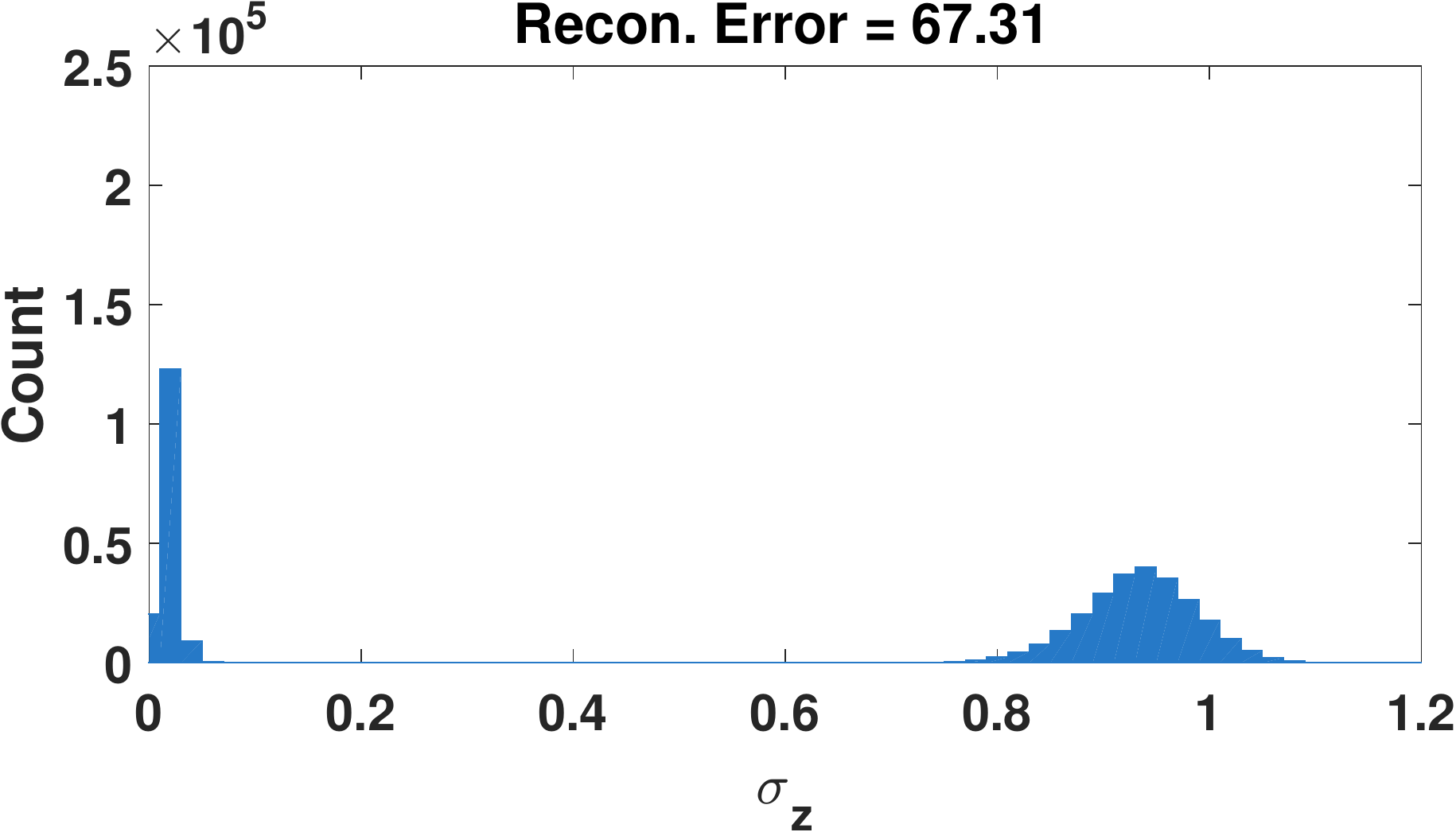}
    \caption{Histogram of $\bsigma_z$ values as VAE encoder/decoder network depth is varied. There are $2$, $4$ and $5$ convolution layers in each spatial scale from left to right. As depth increases, the reconstruction error grows and more $\bsigma_z$ values are near $1$, indicative of impending posterior collapse.}
    \label{fig:hist_sd_z}
\end{figure}

To help corroborate this association between bad AE local minima and VAE posterior collapse, we plot histograms of VAE $\bsigma_z$ values as network depth is varied in Figure~\ref{fig:hist_sd_z}. The models are trained on CelebA and the number of convolution layers in each spatial scale is $2$, $4$ and $5$ from left to right. As the depth increases, the reconstruction error becomes larger and there are more $\bsigma_z$ near $1$.

\section{Discussion} \label{sec:conclusions}
In this work we have emphasized the previously-underappreciated role of bad local minima in trapping VAE models at posterior collapsed solutions.  Unlike affine decoder models whereby all local minima are provably global, Proposition \ref{prop:non_affine_special_case} stipulates that even infinitesimal nonlinear perturbations can introduce suboptimal local minima characterized by deleterious posterior collapse.  Furthermore, we have demonstrated that the risk of converging to such a suboptimal minima increases with decoder depth.  In particular, we outline the following practically-likely pathway to posterior collapse:
\begin{enumerate}[leftmargin=*]
\item Deeper AE architectures are essential for modeling high-fidelity images or similar, and yet counter-intuitively, increasing AE depth can actually produce larger reconstruction errors on the training data because of bad local minima (with or without skip connections).  An analogous VAE model with the same architecture will likely produce even worse reconstructions because of the additional KL regularization term, which is not designed to steer optimization trajectories away from poor reconstructions.

\item At any such bad local minima, the value of $\gamma$ will necessarily be large, i.e., if it is not large, we cannot be at a local minimum.

\item But because of the thresholding behavior of the VAE as quantified by Proposition \ref{prop:guaranteed_posterior_collapse}, as $\gamma$ becomes larger there is an increased risk of exact posterior collapse along excessive latent dimensions.  And complete collapse along all dimensions will occur for some finite $\gamma$ sufficiently large. Furthermore, explicitly forcing $\gamma$ to be small does not fix this problem, since in some sense the \emph{implicit} $\gamma^*$ is still large as discussed in Section \ref{sec:practical_implications}.


\end{enumerate}

While we believe that this message is interesting in and of itself, there are nonetheless several practically-relevant implications.  For example, complex hierarchical VAEs like BIVA notwithstanding, skip connections and KL warm-start have modest ability to steer optimization trajectories towards good solutions; however, this underappreciated limitation will not generally manifest until networks are sufficiently deep as we have considered.  Fortunately, any advances or insights gleaned from developing deeper unregularized AEs, e.g., better AE architectures, training procedures, or initializations \citep{li2019random}, could likely be adapted to reduce the risk of posterior collapse in corresponding VAE models.

In closing, we should also mention that, although this work has focused on Gaussian VAE models, many of the insights translate into broader non-Gaussian regimes.  For example, a variety of recent VAE enhancements involve replacing the fixed Gaussian latent-space prior $p(\bz)$ with a parameterized non-Gaussian alternative \citep{bauer2019resampled,tomczak2018vae}.  This type of modification provides greater flexibility in modeling the aggregated posterior in the latent space, which is useful for generating better samples \citep{makhzani2016}.  However, it does not immunize VAEs against the bad local minima introduced by deep decoders, and good reconstructions are required by models using Gaussian or non-Gaussian priors alike.  Therefore, our analysis herein still applies in much the same way.

%
%




\begin{appendix}

\section{Network Structure, Experimental Settings, and Additional ImageNet Results} \label{sec:experimental_details}

Three different kinds of network structures are used in the experiments: fully connected networks, convolution networks, and residual networks. For all these structures, we set the dimension of the latent variable $\bz$ to $64$. We then describe the network details accordingly.

\textbf{Fully Connected Netowrk:} This experiment is only applied on the simple Fashion-MNIST dataset, which contains $60000$ $28\times28$ black-and-while images. These images are first flattened to a $784$ dimensional vector. Both the encoder and decoder have multiple number of $512$-dimensional hidden layers, each followed by ReLU activations.

\textbf{Convolution Netowrk:} The original images are either $32\times32\times3$ (Cifar10, Cifar100 and SVHN) or $64\times64\times3$ (CelebA and ImageNet). In the encoder, we use a multiple number (denoted as $t$) of $3\times3$ convolution layers for each spatial scale. Each convolution layer is followed by a ReLU activation. Then we use a $2\times2$ max pooling to downsample the feature map to a smaller spatial scale. The number of channels is doubled when the spatial scale is halved. We use $64$ channels when the spatial scale is $32\times32$. When the spatial scale reaches $4\times4$ (there should be $512$ channels in this feature map), we use an average pooling to transform the feature map to a vector, which is then transformed into the latent variable using a fully connected layer. In the decoder, the latent variable is first transformed to a $4096$-dimensional vector using a fully connected layer and then reshaped to $2\times2\times1024$. Again in each spatial scale, we use $1$ transpose convolution layer to upscale the feature map and halve the number of channels followed by $t-1$ convolution layers. Each convolution and transpose convolution layer is followed by a ReLU activation layer. When the spatial scale reaches that of the original image, we use a convolution layer to transofrm the feature map to $3$ channels.

\textbf{Residual Network:} The network structure of the residual network is similar to that of a convolution network described above. We simply replace the convolution layer with a residual block. Inside the residual block, we use different numbers of convolution numbers. (The typical number of convolution layers inside a residual block is $2$ or $3$. In our experiments, we try $2$, $3$, $4$ and $5$.)

\textbf{Training Details:} All the experiments with different network structures and datasets are trained in the same procedure. We use the Adam optimization method and the default optimizer hyper parameters in Tensorflow. The batch size is $64$ and we train the model for $250K$ iterations. The initial learning rate is $0.0002$ and it is halved every $100K$ iterations.

\begin{figure}[h]
    \centering
    \includegraphics[width=0.5\linewidth]{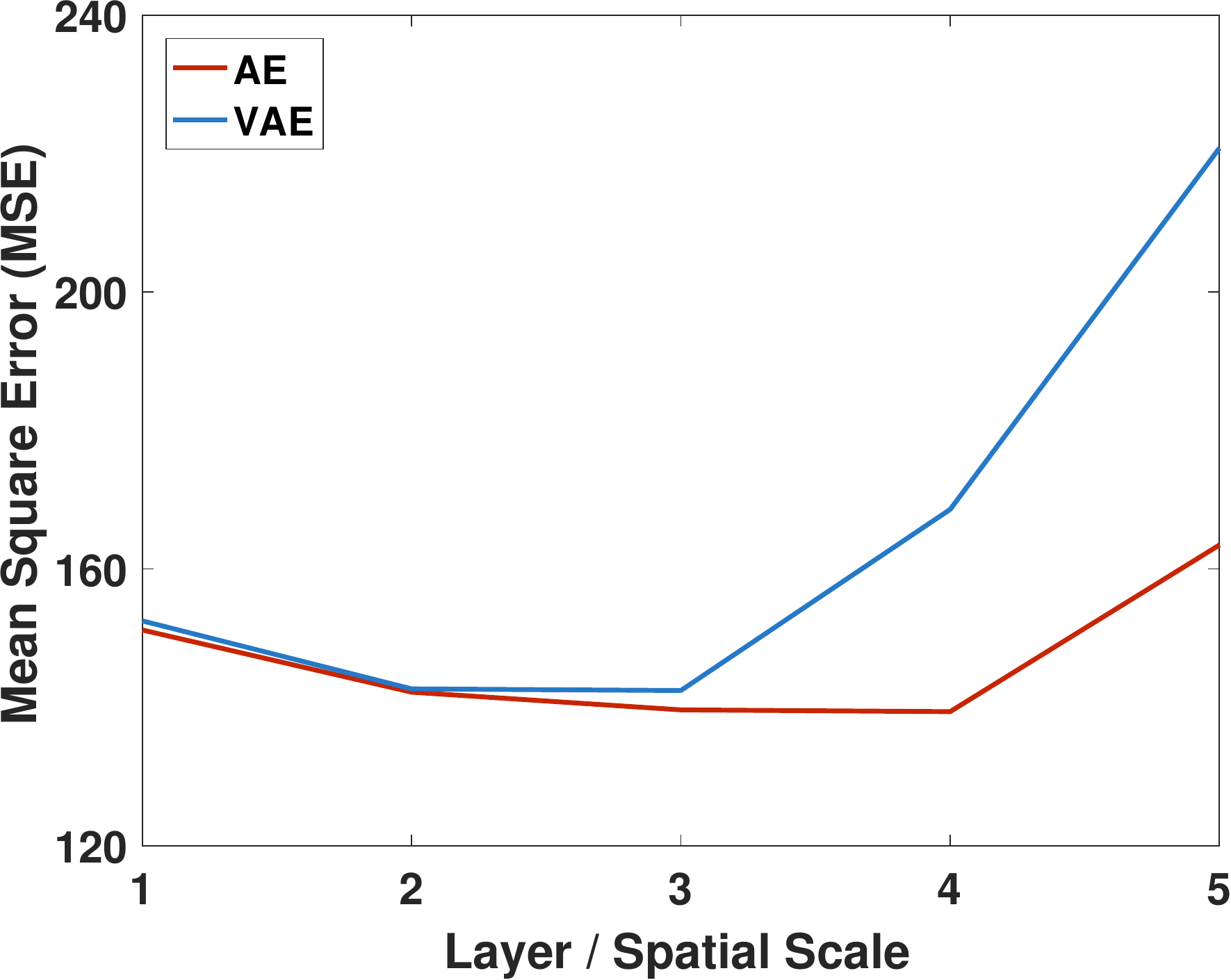}
    \caption{Reconstruction error for Convolution networks with increasing depth/\# of spatial scales trained on ImageNet.}
    \label{fig:recon_imagenet}
\end{figure}

\textbf{Additional Results on ImageNet:} We also show the reconstruction error for convolution networks with increasing depth trained on ImageNet in Figure~\ref{fig:recon_imagenet}. The trend is the same as that in Figure~\ref{fig:recon_err}.

\section{Proof of Proposition \ref{prop:non_affine_special_case}}

\newcommand{\data}[1]{\bx^{(#1)}}
\newcommand{\partialderiv}[2]{\frac{\partial #2}{\partial #1}}
\newcommand{\ppartialderiv}[3]{\frac{\partial^2 #3}{\partial #1 #2}}

While the following analysis could in principle be extended to more complex datasets, for our purposes it is sufficient to consider the following simplified case for ease of exposition.  Specifically, we assume that $n>1,d>\kappa$, set $d=2,n=2,\kappa=1$, and $\data{1}=(1,1), \data{2}=(-1,-1)$.

Additionally, we will use the following basic facts about the Gaussian tail. Note that \eqref{eq:suppl-trunc-normal-bound1}-\eqref{eq:suppl-trunc-normal-bound2} below follow from integration by parts; see \cite{Orjebin14}.
\begin{lemma}
    Let $\epsilon\sim \calN(0,1),A>0$; $\phi(x),\Phi(x)$ be the pdf and cdf of the standard normal distribution, respectively. Then
    \begin{align}
        1-\Phi(A) &\le e^{-A^2/2},\label{eq:suppl-trunc-normal-bound0}\\
        \mathbb{E}[\epsilon\mathbf{1}_{\{\epsilon>A\}}] &= \phi(A)
        ,\label{eq:suppl-trunc-normal-bound1}\\
        \mathbb{E}[\epsilon^2\mathbf{1}_{\{\epsilon>A\}}] &= 1-\Phi(A)+A\phi(A)
        .\label{eq:suppl-trunc-normal-bound2}
    \end{align}
\end{lemma}

\subsection{Suboptimality of \eqref{eq:non_affine_bad_local_minimum}}

Under the specificed conditions, the energy from \eqref{eq:non_affine_bad_local_minimum} has a value of $nd$. Thus to show that it is not the global minimum, it suffices to show that the following VAE, parameterized by $\delta$, has energy $\rightarrow-\infty$ as $\delta\rightarrow 0$:
\begin{align*}
\mu_z^{(1)} &= 1, \mu_z^{(2)} = - 1, \\
\bW_x &= (\alpha + 1, \alpha + 1),
 \bb_x = 0, \\
\sigma_z^{(1)} &= \sigma_z^{(2)} = \delta, \\
\gamma &= \mathbb{E}_{\calN (\varepsilon |0, 1)} 2 (1 -
   \pi_{\alpha} ((\alpha + 1) (1 + \delta \varepsilon)))^2.
\end{align*}
This follows because, given the stated parameters, we have that
\begin{align}
  \calL(\theta,\phi)  = & \sum_{i = 1}^2 (
      1 + 2 \log \mathbb{E}_{\calN(\varepsilon |0, 1)} 2 (1 - \pi_{\alpha} ((\alpha + 1) (1 + \delta \varepsilon)))^2  
      - 2 \log \delta + \delta^2 
       + 1) 
        \nonumber\\
   = & \sum_{i = 1}^2 (\Theta (1) + 2 \log \mathbb{E}_{\calN
  (\varepsilon |0, 1)} (1 - \pi_{\alpha} (\alpha + 1 + (\alpha + 1) \delta
  \varepsilon))^2 - 2 \log \delta) \nonumber\\
   \leq^{(i)} & 4 \log \delta + \Theta (1) \nonumber.
\end{align}

(i) holds when $\delta < \frac{1}{\alpha + 1}$; to see this, denote $x
:= \alpha + 1 + (\alpha + 1) (\delta \varepsilon)$. Then
\begin{align*}
   & \mathbb{E}_{\calN (\varepsilon |0, 1)} (1 - \pi_{\alpha}
  (x))^2\\
   = & \mathbb{E}_{\varepsilon} [(1 - \pi_{\alpha} (x))^2
  \mathbf{1}_{\{ x \geq \alpha \}}] +\mathbb{E}_{\varepsilon} [(1 -
  \pi_{\alpha} (x))^2 \mathbf{1}_{\{ | x | < \alpha \}}]
  +\mathbb{E}_{\varepsilon} [(1 - \pi_{\alpha} (x))^2 \mathbf{1}_{\{ x <
  - \alpha \}}]\\
   \leq & \underbrace{\mathbb{E}_{\varepsilon} [(1 - (x -
  \alpha))^2]}_{(a)} + \underbrace{\mathbb{P} (| x | < \alpha)}_{(b)} +
  \underbrace{\mathbb{E}_{\varepsilon} ((1 - x - \alpha)^2 \mathbf{1}_{\{ x
  < - \alpha \}})}_{(c)} .
\end{align*}
In the RHS above $(a) = [(\alpha + 1) \delta]^2$; using \eqref{eq:suppl-trunc-normal-bound0}-\eqref{eq:suppl-trunc-normal-bound2} we then have
\begin{align*}
(b) &<\mathbb{P} (x < \alpha) =\mathbb{P} \left( \varepsilon < \frac{-
   1}{(\alpha + 1) \delta} \right) \leq \exp \left( - \frac{1}{2 [(\alpha + 1)
   \delta]^2} \right) .\\
  (c) & <  \mathbb{E}_{\varepsilon} ((2 \alpha + (\alpha + 1) \delta
  \varepsilon)^2 \mathbf{1}_{\{ x < \alpha \}}) \\
  & =  \int_{- \infty}^{\frac{- 1}{(\alpha + 1) \delta}} (2 \alpha + (\alpha
  + 1) \delta \varepsilon)^2  \frac{1}{\sqrt{2 \pi}} e^{- \varepsilon^2 / 2} d
  \varepsilon \\
  & <  \int_{- \infty}^{\frac{- 1}{(\alpha + 1) \delta}} (4 \alpha^2 +
  [(\alpha + 1) \delta \varepsilon]^2)  \frac{1}{\sqrt{2 \pi}} e^{-
  \varepsilon^2 / 2} d \varepsilon \\ 
  & <  \left\{ 4 \alpha^2 + ((\alpha + 1) \delta)^2 \left[ 1 +
  \frac{1}{\sqrt{2 \pi}} \right] \right\} \exp \left( - \frac{1}{2 [(\alpha +
  1) \delta]^2} \right)
\end{align*}
when $\delta < \frac{1}{\alpha + 1}$. 
Thus
$$
  \lim_{\delta \rightarrow 0}  \frac{\mathbb{E}_{\calN (\varepsilon |0,
  1)} (1 - \pi_{\alpha} (x))^2}{[(\alpha + 1) \delta]^2} = 1,
$$
and
$$
\lim_{\delta \rightarrow 0} \{ \log \mathbb{E}_{\calN (\varepsilon
   |0, 1)} (1 - \pi_{\alpha} (
   x))^2 - 2 \log \delta \} = 2 \log (\alpha + 1),
$$
or
$$
2\log \mathbb{E}_\epsilon(1-\pi_\alpha(x))^2=4\log \delta+\Theta(1),
$$
and we can see (i) holds.

\subsection{Local Optimality of \eqref{eq:non_affine_bad_local_minimum}}

We will now show that at \eqref{eq:non_affine_bad_local_minimum}, the Hessian of the energy has structure
$$
\begin{array}{ccccc}
& (\bW_x) & (\bb_x) & (\sigma^{(i)}_z,\mu_z^{(i)}) & (\gamma)\\
(\bW_x)           & 0 & 0 & 0 & 0 \\
(\bb_x)           & 0 & \frac{2}{\gamma}I & 0 & 0 \\
(\sigma^{(i)}_z,\mu_z^{(i)}) & 0 & 0 & (p.d.) & 0 \\
(\gamma)       & 0 & 0 & 0 & (p.d.)
\end{array}
$$
where p.d. means the corresponding submatrix is positive definite and independent of other parameters.
While the Hessian is $0$ in the subspace of $\bW_x$, we can show that for VAEs that are only different from \eqref{eq:non_affine_bad_local_minimum} by $\bW_x$, the gradient always points back to \eqref{eq:non_affine_bad_local_minimum}. Thus \eqref{eq:non_affine_bad_local_minimum} is a strict local minima.

First we compute the Hessian matrix block-wise.
We will identify $\bW_x\in\mathbb{R}^{2\times 1}$ with the vector $(W_j)_{j=1}^2$, and use the shorthand notations $\bx^{(i)}=(x^{(i)}_j)_{j=1}^2$, $\bb_x=(b_j)_{j=1}^2$, $z^{(i)} = \mu_z^{(i)} + \sigma_z^{(i)} \varepsilon$, where $\varepsilon\sim \calN(0,1)$ (recall that $z^{(i)}$ is a scalar in this proof).
\begin{enumerate}
    \item The second-order derivatives involving $\bW_x$ can be expressed as
    \begin{align}
    \partialderiv{W_j} \calL &=  \frac{-2}{\gamma} \sum_{i=1}^n
       \mathbb{E}_{\varepsilon}[
            (\pi_{\alpha}'(W_j z^{(i)}) z^{(i)}) \cdot (x^{(i)}_j - \pi_{\alpha} (W_j z^{(i)}) - b_j)
        ], \label{eq:supp-non-affine-grad-W}
    \end{align}
    and therefore all second-order derivatives involving $W_j$ will have the form
    \begin{equation}\label{eq:supp-non-affine-w-hessian-form}
    \mathbb{E}_\epsilon[\pi'_\alpha(W_j z^{(i)})F_1 +
                        \pi''_\alpha(W_j z^{(i)})F_2],
    \end{equation}
    where $F_1,F_2$ are some arbitrary functions that are finite at \eqref{eq:non_affine_bad_local_minimum}.
    Since $\pi'_\alpha(0)=\pi''_\alpha(0)=W_j=0$, the above always evaluates to $0$ at $\bW_x=0$.
    \item For second-order derivatives involving $\bb_x$, we have
    $$
    \partialderiv{\bb_x}{\calL} =
        \frac{-2}{\gamma} \mathbb{E}_{\varepsilon}[
            \bx^{(i)} - \pi_{\alpha} (\bW_x z^{(i)}) - \bb_x]
    $$
    and
    \begin{align*}
        \frac{\partial^2\calL}{\partial (\bb_x)^2} &= \frac{2}{\gamma}I, \\
        \frac{\partial^2\calL}{\partial\gamma \partial\bb_x} &= \frac{2}{\gamma^2}\partialderiv{\bb_x}{\calL} = 0, &(\text{since }\bW_x=0);
    \end{align*}
    and $\frac{\partial^2\calL}{\partial\mu_z^{(i)} \partial\bb_x}$ and $\frac{\partial^2\calL}{\partial\mu_z^{(i)} \partial\sigma_z^{(i)}}$ will also have the form of \eqref{eq:supp-non-affine-w-hessian-form}, thus both equal $0$ at $\bW_x=0$.

    \item Next consider second-order derivatives involving $\mu_z^{(i)}$ or $\sigma^{(i)}_k$. Since the KL part of the energy, $\sum_{i=1}^n\mathrm{KL}(q_\phi(\bz|\bx^{(i)})|p(\bz))$, only depends on $\mu_z^{(i)}$ and $\sigma^{(i)}_k$, and have p.d. Hessian at \eqref{eq:non_affine_bad_local_minimum} independent of other parameters, it suffices to calculate the derivatives of the reconstruction error part, denoted as $\calL_\text{recon}$. Since
    \begin{align*} 
    \partialderiv{\mu_z^{(i)}}{\calL_\text{recon}} &=
    \frac{-2}{\gamma}\sum_{i,j}\mathbb{E}_\epsilon\left[
        (x_j^{(i)}-\pi_\alpha(W_j z^{(i)})-b_j)
        W_j \pi'_\alpha(W_j z^{(i)})
    \right],\\
    \partialderiv{\sigma_z^{(i)}}{\calL_\text{recon}} &=
    \frac{-2}{\gamma}\sum_{i,j}\mathbb{E}_\epsilon\left[
        (x_j^{(i)}-\pi_\alpha(W_j z^{(i)})-b_j)
        W_j \epsilon \pi'_\alpha(W_j z^{(i)})
    \right],
    \end{align*}
    all second-order derivatives will have the form of \eqref{eq:supp-non-affine-w-hessian-form}, and equal $0$ at $\bW_x=0$.

    \item For $\gamma$, we can calculate that $\partial^2\calL/\partial\gamma^2=4/\gamma^2>0$ at \eqref{eq:non_affine_bad_local_minimum}.
\end{enumerate}

Now, consider VAE parameters that are only different from \eqref{eq:non_affine_bad_local_minimum} in $\bW_x$. Plugging $\bb_x=\bar{\bx},\mu_z^{(i)}=0,\sigma^{(i)}_k=1$ into \eqref{eq:supp-non-affine-grad-W}, we have

\begin{align*}
    \partialderiv{W_j} \calL &= \frac{-2}{\gamma} \sum_{i=1}^n
       \mathbb{E}_{\varepsilon}[
            (\pi_{\alpha}'(W_j \varepsilon) \varepsilon) \cdot (-\pi_{\alpha} (W_j \varepsilon))
        ].
\end{align*}

As $(\pi_{\alpha}'(W_j \varepsilon) \varepsilon) \cdot (-\pi_{\alpha} (W_j \varepsilon))\le 0$ always holds, we can see that the gradient points back to \eqref{eq:non_affine_bad_local_minimum}. This concludes our proof of \eqref{eq:non_affine_bad_local_minimum} being a strict local minima.  \myendofproof

\section{Proof of Proposition \ref{prop:guaranteed_posterior_collapse}}

We begin by assuming an arbitrarily complex encoder for convenience.  This allows us to remove the encoder-sponsored amortized inference and instead optimize independent parameters $\bmu_z^{(i)}$ and $\bsigma_z^{(i)}$ separately for each data point.  Later we will show that this capacity assumption can be dropped and the main result still holds.

We next define
\begin{equation}
\bm_z \triangleq \left[\left( \bmu_z^{(1)} \right)^{\top}, \ldots, \left( \bmu_z^{(n)} \right)^{\top}     \right]^{\top} \in \mathbb{R}^{\kappa n}  ~~ \mbox{and} ~~ \bs_z \triangleq \left[\left( \bsigma_z^{(1)} \right)^{\top}, \ldots, \left( \bsigma_z^{(n)} \right)^{\top}     \right]^{\top} \in \mathbb{R}^{\kappa n},
\end{equation}
which are nothing more than the concatenation of all of the decoder means and variances from each data point into the respective column vectors.  It is also useful to decompose the assumed non-degenerate decoder parameters via
\begin{equation}
\theta ~ \equiv ~ \left[ \psi, w \right], ~~~ \psi ~ \triangleq ~ \theta \backslash w,
\end{equation}
where $w \in [0,1]$ is a scalar such that $\mu_x\left(\bz; \theta \right) \equiv \mu_x\left(w \bz; \psi \right)$.  Note that we can always reparameterize an existing deep architecture to extract such a latent scaling factor which we can then hypothetically optimize separately while holding the remaining parameters $\psi$ fixed.  Finally, with slight abuse of notation, we may then define the function
\begin{eqnarray}
\hspace*{0.1cm} f\left(w \bm_z, w \bs_z \right) ~ \triangleq  && \\
&& \hspace*{-2.5cm} \sum_{i=1}^n f\left(\bmu_z^{(i)}, \bsigma_z^{(i)}, [\tilde{\psi}, w],\bx^{(i)}\right) \equiv \sum_{i=1}^n \mathbb{E}_{\calN \left( \bz | \bmu_z^{(i)}, \mbox{diag}\left[\bsigma_z^{(i)}\right]^2 \right)} \left[ \|\bx^{(i)} - \bmu_x \left(w \bz; \tilde{\psi} \right)  \|_2^2  \right]. \nonumber
\end{eqnarray}
This is basically just the original function $f$ summed over all training points, with $\psi$ fixed at the corresponding values extracted from $\tilde{\theta}$ while $w$ serves as a free scaling parameter on the decoder.



Based on the assumption of Lipschitz continuous gradients, we can always create the upper bound
\begin{eqnarray}
\hspace*{-0.0cm} f\left(\bu,\bv  \right) ~~ \leq ~~ f\left(\tilde{\bu},\tilde{\bv} \right) && \\
&& \hspace*{-4.0cm} ~+~ \left(\bu  - \tilde{\bu} \right)^{\top} \left. \nabla_{u } f\left(\bu ,\bv   \right) \right|_{\bu = \tilde{\bu}} ~+~ \tfrac{L}{2} \left\|\bu  - \tilde{\bu}  \right\|_2^2 ~+~ \left(\bv  - \tilde{\bv} \right)^{\top}\left. \nabla_{v} f\left(\bu ,\bv   \right) \right|_{\bv = \tilde{\bv}} ~+~ \tfrac{L}{2} \left\|\bv  - \tilde{\bv}  \right\|_2^2, \nonumber
\end{eqnarray}
where $L$ is the Lipschitz constant of the gradients and we have adopted $\bu \triangleq w \bm_z$ and $\bv \triangleq w \bsigma_z$ to simplify notation.  Equality occurs at the evaluation point $\{ \bu,\bv \}  = \{ \tilde{\bu}, \tilde{\bv} \}$.  However, this bound does not account for the fact that we know $\nabla_{v} f\left(\bu ,\bv   \right)  \geq 0$ (i.e., $f\left(\bu ,\bv   \right)$ is increasing w.r.t. $\bv$) and that $\bv \geq 0$.  Given these assumptions, we can produce the refined upper bound
\begin{equation}
f^{ub}\left(\bu,\bv \right) ~ \geq ~ f\left(\bu,\bv  \right),
\end{equation}
where ~~ $f^{ub}\left(\bu,\bv  \right) ~ \triangleq$
\vspace*{-0.2cm}
\begin{equation}
f\left(\tilde{\bu},\tilde{\bv} \right) ~+~ \left(\bu  - \tilde{\bu} \right)^{\top}\left. \nabla_{u } f\left(\bu ,\bv   \right) \right|_{\bu = \tilde{\bu}} ~+~ \tfrac{L}{2} \left\|\bu  - \tilde{\bu}  \right\|_2^2 + \sum_{j=1}^{nd} g\left(v_j , \tilde{v}_j,  \left.\nabla_{v_j} f\left(\bu ,\bv   \right) \right|_{v_j = \tilde{v_j}} \right)
\end{equation}
and the function $g : \mathbb{R}^3 \rightarrow \mathbb{R}$ is defined as
\begin{equation}
g\left(v , \tilde{v}, \delta \right) \triangleq  \left\{\begin{array}{cc} \left(v  - \tilde{v} \right) \delta + \tfrac{L}{2} \left(v  - \tilde{v}  \right)_2^2 \vspace*{0.2cm} & \mbox{if} ~~ v \geq \tilde{v} - \tfrac{\delta}{L} ~~ \mbox{and} ~~ \{v,\tilde{v},\delta\} \geq 0, \\ \tfrac{-\delta^2}{2L} \vspace*{0.2cm} & \mbox{if} ~~ v < \tilde{v} - \tfrac{\delta}{L} ~~ \mbox{and} ~~ \{v,\tilde{v},\delta\} \geq 0, \\ \infty & \mbox{otherwise}. \end{array}   \right.
\end{equation}
Given that
\begin{equation}
\tilde{v} - \tfrac{\delta}{L} = \arg \min_v \left[ \left(v  - \tilde{v} \right) \delta + \tfrac{L}{2} \left(v  - \tilde{v}  \right)_2^2 \right] ~~ \mbox{and} ~~ \tfrac{-\delta^2}{2L} = \min_v \left[ \left(v  - \tilde{v} \right) \delta + \tfrac{L}{2} \left(v  - \tilde{v}  \right)_2^2 \right],
\end{equation}
the function $g$ is basically just setting all values of $\left(v  - \tilde{v} \right) \delta + \tfrac{L}{2} \left\|v  - \tilde{v}  \right\|_2^2$ with negative slope to the minimum $\tfrac{-\delta^2}{2L}$.  This change is possible while retaining an upper bound because $f\left(\bu ,\bv   \right)$ is non-decreasing in $\bv$ by stated assumption. Additionally, $g$ is set to infinity for all $v < 0$ to enforce non-negatively.

While it may be possible to proceed further using $f^{ub}$, we find it useful to consider a final modification.  Specifically, we define the approximation
\begin{equation}
f^{appr}\left(\bu,\bv \right)  ~~ \approx ~~ f^{ub}\left(\tilde{\bu},\tilde{\bv} \right),
\end{equation}
where ~~ $f^{appr}\left(\bu,\bv  \right) ~ \triangleq$
\vspace*{-0.2cm}
\begin{equation}
f\left(\tilde{\bu},\tilde{\bv} \right) ~+~ \left(\bu  - \tilde{\bu} \right)^{\top} \left. \nabla_{u } f\left(\bu ,\bv   \right) \right|_{\bu = \tilde{\bu}} ~+~ \tfrac{L}{2} \left\|\bu  - \tilde{\bu}  \right\|_2^2 + \sum_{j=1}^{nd} g^{appr} \left(v_j , \tilde{v}_j,  \left.\nabla_{v_j} f\left(\bu ,\bv   \right) \right|_{v_j = \tilde{v_j}} \right)
\end{equation}
and
\begin{equation} \label{eq:approximate_g_function}
g^{appr}\left(v , \tilde{v}, \delta \right) \triangleq  \left\{\begin{array}{cc} \tfrac{-\delta^2}{2L} + \tfrac{\delta^2}{2L\tilde{v}^2}v^2 \vspace*{0.2cm} & \mbox{if} ~~ \tilde{v} - \tfrac{\delta}{L} \geq 0 ~~ \mbox{and} ~~ \{v,\tilde{v},\delta\} \geq 0, \\ \left( \tfrac{L\tilde{v}^2}{2} - \delta \tilde{v} \right)  + \left(\tfrac{\delta}{\tilde{v}} - \tfrac{L}{2}  \right) v^2 \vspace*{0.2cm} & \mbox{if} ~~ \tilde{v} - \tfrac{\delta}{L} < 0 ~~ \mbox{and} ~~ \{v,\tilde{v},\delta\} \geq 0, \\ \infty & \mbox{otherwise}. \end{array}   \right.
\end{equation}
While slightly cumbersome to write out, $g^{appr}$ has a simple interpretation. By construction, we have that
\begin{equation} \label{eq:equivalence_points}
\min_v g^{appr}\left(v , \tilde{v}, \delta \right) = g^{appr}\left(0 , \tilde{v}, \delta \right) = \min_v g\left(v , \tilde{v}, \delta \right) = g\left(0 , \tilde{v}, \delta \right)
\end{equation}
\vspace*{-0.2cm}
and
\vspace*{-0.1cm}
\begin{equation} \label{eq:equivalence_points2}
g^{appr}\left(\tilde{v} , \tilde{v}, \delta \right) = g\left(\tilde{v} , \tilde{v}, \delta \right) = 0.
\end{equation}
At other points, $g^{appr}$ is just a simple quadratic interpolation but without any factor that is linear in $v$.  And removal of this linear term, while retaining (\ref{eq:equivalence_points}) and  (\ref{eq:equivalence_points}) will be useful for the analysis that follows below.  Note also that although $f^{appr}\left(\bu,\bv  \right)$ is no longer a strict bound on $f\left(\bu,\bv  \right)$, it will nonetheless still be an upper bound whenever $v_j \in \{0,\tilde{v}_j\}$ for all $j$ which will ultimately be sufficient for our purposes.

We now consider optimizing the function
\begin{equation}
h^{appr}(\bm_z,\bs_z,w) \triangleq \tfrac{1}{\gamma} f^{appr}\left(w \bm_z,w \bs_z \right) + \sum_{i=1}^n \left\| \bmu_z^{(i)} \right\|_2^2 + \left\| \bsigma_z^{(i)} \right \|_2^2 - \log \left| \mbox{diag}\left[\bsigma_z^{(i)}  \right]^2 \right|.
\end{equation}
If we define $\calL \left(\bm_z, \bs_z, w \right)$ as the VAE cost from (\ref{eq:VAE_objective_general_Gaussian}) under the current parameterization, then by design it follows that
\begin{equation}
h^{appr}(\tilde{\bm}_z,\tilde{\bs}_z,\tilde{w}) = \calL \left(\tilde{\bm}_z,\tilde{\bs}_z,\tilde{w} \right)
\end{equation}
and
\begin{equation}
h^{appr}(\bm_z,\bs_z,w) \geq \calL \left( \bm_z,\bs_z,w \right)
\end{equation}
whenever $w \sigma_j \in \{0, \tilde{w} \tilde{\sigma}_j\}$ for all $j$.  Therefore if we find such a solution $\{ \bm'_z,\bs'_z,w' \}$ that satisfies this condition and has $h^{appr}(\bm'_z,\bs'_z,w') < h^{appr}(\tilde{\bm}_z,\tilde{\bs}_z,\tilde{w})$, it necessitates that $\calL(\bm'_z,\bs'_z,w') < \calL(\tilde{\bm}_z,\tilde{\bs}_z,\tilde{w})$ as well.  This then ensures that $\{ \tilde{\bm}_z,\tilde{\bs}_z,\tilde{w} \}$ cannot be a local minimum.

We now examine the function $h^{appr}$ more closely.  After a few algebraic manipulations and excluding irrelevant constants, we have that
\begin{eqnarray}
\hspace*{-0.5cm} h^{appr}(\bm_z,\bs_z,w) \equiv && \nonumber \\
&& \hspace*{-3cm} \sum_{j=1}^{nd} \left\{ \tfrac{1}{\gamma}\left[  w m_{z,j} \left. \nabla_{u_j } f\left(\bu ,\bv   \right) \right|_{u_j = \tilde{w} \tilde{m}_{z,j}} + \tfrac{L}{2} \left(w^2 m_{z,j}^2 - 2 w m_{z,j} \tilde{w} \tilde{m}_{z,j} \right) + c_j w^2 s_{z,j}^2 \right] \right. \nonumber \\
&& \hspace*{-2.2cm}  \left. + ~~  m_{z,j}^2 + s_{z,j}^2 - \log s_{z,j}^2 \right\},
\end{eqnarray}
where $c_j$ is the coefficient on the $v^2$ term from (\ref{eq:approximate_g_function}).  After rearranging terms, optimizing out $\bm_z$ and $\bs_z$, and discarding constants, we can then obtain (with slight abuse of notation) the reduced function
\begin{equation} \label{eq:h_function_small_form}
h^{appr}(w) \triangleq \sum_{j=1}^{nd} \frac{y_j}{\gamma + \beta  w^2} + \log(\gamma + c_j w^2),
\end{equation}
where $\beta \triangleq \tfrac{L}{2}$ and $y_j \triangleq \tfrac{L}{2} \left\| \tilde{w} \tilde{m}_{z,j} - \tfrac{1}{L} \left. \nabla_{u_j } f\left(\bu ,\bv   \right) \right|_{u_j = \tilde{w} \tilde{m}_{z,j}} \right\|_2^2$.  Note that $y_j$ must be bounded since $L \neq 0$\footnote{$L=0$ would violate the stipulated conditions for a non-degenerate decoder since it would imply that no signal from $\bz$ could pass through the decoder.  And of course if $L = 0$, we would already be at a solution exhibiting posterior collapse.} and $w \in [0,1]$, $\left. \nabla_{u_j } f\left(\bu ,\bv   \right) \right|_{u_j = \tilde{w} \tilde{m}_{z,j}} \leq L$, and $\tilde{\bm}$ are all bounded.  The latter is implicitly bounded because the VAE KL term prevents infinite encoder mean functions.  Furthermore, $c_j$ must be strictly greater than zero per the definition of a non-degenerate decoder; this guarantees that
\begin{equation}
g^{appr} \left(\tilde{w} \tilde{s}_j , \tilde{w} \tilde{s}_j,  \left.\nabla_{v_j} f\left(\bu ,\bv   \right) \right|_{v_j = \tilde{w} \tilde{s_j}} \right) > g^{appr} \left(0 , \tilde{w} \tilde{s}_j,  \left.\nabla_{v_j} f\left(\bu ,\bv   \right) \right|_{v_j = \tilde{w} \tilde{s_j}} \right),
\end{equation}
which is only possible with $c_j > 0$.  Proceeding further, because
\begin{equation}
\nabla_{w^2} h^{appr}(w) = \sum_{j=1}^{nd} \left( \frac{-\beta y_j}{\left(\gamma + \beta  w^2\right)^2} + \frac{c_j}{\gamma + c_j w^2} \right),
\end{equation}
we observe that if $\gamma$ is increased sufficiently large, the first term will always be smaller than the second since $\beta$ and all $y_j$ are bounded, and $c_j > 0$ $\forall j$.   So there can never be a point whereby $\nabla_{w^2} h^{appr}(w) = 0$ when $\gamma = \gamma'$ sufficiently large.  Therefore the minimum in this situation occurs on the boundary where $w^2 = 0$.  And finally, if $w^2 = 0$, then the optimal $\bm_z$ and $\bs_z$ is determined solely by the KL term, and hence they are set according to the prior.  Moreover, the decoder has no signal from the encoder and is therefore optimized by simply setting $\bmu_x\left( 0 ; \tilde{\psi}\right)$ to the mean $\bar{\bx}$ for all $i$.\footnote{We are assuming here that the decoder has sufficient capacity to model any constant value, e.g., the output layer has a bias term.}  Additionally, none of this analysis requires and arbitrarily complex encoder; the exact same results hold as long as the encoder can output a 0 for means and 1 for the variances.

Note also that if we proceed through the above analysis using $\bw \in \mathbb{R}^{\kappa}$ as parameterizing a separate $w_j$ scaling factor for each latent dimension $j \in \{1,\ldots,\kappa\}$, then a smaller $\gamma$ value would generally force partial collapse.  In other words, we could enforce nonzero gradients of $h^{appr}(w)$ along the indices of each latent dimension separately.  This loosely criteria would then lead to $q_{\phi^*}(z_j | \bx) = p(z_j)$ along some but not all latent dimensions as stated in the main text below Proposition \ref{prop:guaranteed_posterior_collapse}.  \myendofproof

\section{Representative Stationary Point Exhibiting Posterior Collapse in Deep VAE Models}
\label{sec:stationary_point_example}

Here we provide an example of a stationary point that exhibits posterior collapse with an arbitrary deep encoder/decoder architecture. This example is representative of many other possible cases.  Assume both encoder and decoder mean functions $\bmu_x$ and $\bmu_z$, as well as the diagonal encoder covariance function $\bSigma_z = \mbox{diag}[\bsigma_z^2]$, are computed by standard deep neural networks, with layers composed of linear weights followed by element-wise nonlinear activations (the decoder covariance satisfies $\bSigma_x = \gamma \bI$ as before).  We denote the weight matrix from the first layer of the decoder mean network as $\bW_{\mu_x}^1$, while $\bw^1_{\mu_x,\cdot j}$ refers to the corresponding $j$-th column.  Assuming $\rho$ layers, we denote $\bW^\rho_{\mu_z}$ and $\bW^\rho_{\sigma^2_z}$ as weights from the last layers of the encoder networks producing $\bmu_z$ and $\log \bsigma^2_z$ respectively, with $j$-th rows defined as $\bw^\rho_{\mu_z, j \cdot}$ and $\bw^\rho_{\sigma^2_z, j \cdot}$.  We then characterize the following key stationary point:
\begin{proposition} \label{prop:zero_gradient}
If $\bw^1_{\mu_x,\cdot j} = \left( \bw^\rho_{\mu_z,j\cdot} \right)^{\top} = \left( \bw^\rho_{\sigma^2_z,j\cdot} \right)^{\top} =  {\bf 0}$ for any $j \in \{1,2,\ldots,\kappa\}$, then the gradients of (\ref{eq:VAE_objective_general_Gaussian}) with respect to $\bw^1_{\mu_x,\cdot j}$, $\bw^\rho_{\mu_z,j\cdot}$, and $\bw^\rho_{\sigma^2_z,j\cdot}$  are all equal to zero.
\end{proposition}

If the stated weights are zero along dimension $j$, then obviously it must be that $q_{\phi}(z_j | \bx) = p(z_j)$, i.e., a collapsed dimension for better or worse.  The proof is straightforward; we provide the details below for completeness.


\textbf{\emph{Proof:}}  First we remind that
the variational upper bound is defined in (\ref{eq:VAE_objective}). We define $\mathcal{L}(\bx;\theta,\phi)$ as the loss at a data point $\bx$, \emph{i.e.}
\begin{equation}
	\mathcal{L}(\bx;\theta,\phi) = -\mathbb{E}_{q_\phi(\bz|\bx)} \left[ \log p_\theta(\bx|\bz) \right] + \mathbb{KL}\left[ q_\phi(\bz|\bx) || p(\bz) \right].
\end{equation}
The total loss is the integration of $\mathcal{L}(\bx;\theta,\phi)$ over $\bx$. Further more, we denote $\mathcal{L}_{kl}(\bx;\theta)$ and $\mathcal{L}_{gen}(\bx;\theta,\phi)$ as the KL loss and the generation loss at $\bx$ respectively, \emph{i.e.}
\begin{eqnarray}
	\mathcal{L}_{kl}(\bx;\phi) &=& \mathbb{KL}\left[ q_\phi(\bz|\bx) || p(\bz) \right] = \sum_{i=1}^\kappa \mathbb{KL}\left[ q_\phi(z_j|\bx) || p(z_j) \right], \nonumber \\
	&=& \frac{1}{2}\sum_{j=1}^\kappa \left( \mu_{z,j}^2 + \sigma_{z,j}^2 - \log \sigma_{z,j}^2 - 1 \right) \label{eqn:kl_loss}\\
	\mathcal{L}_{gen}(\bx;\phi,\theta) &=& -\mathbb{E}_{q_\phi(\bz|\bx)} \left[ \log p_\theta(\bx|\bz) \right] . \label{eqn:gen_loss}
\end{eqnarray}
The second equality in (\ref{eqn:kl_loss}) holds because the covariance of $q_\phi(\bz|\bx)$ and $p(\bz)$ are both diagonal. The last encoder layer and the first decoder layer are denoted as $\bh_e^\rho$ and $\bh_d^1$.
If $\bw_{\mu_z,j\cdot}^\rho=0, \bw_{\sigma_z^2,j\cdot}^\rho=0$, then we have
\begin{equation}
\mu_{z,j}=\bw_{\mu_z,j\cdot}^\rho\bh_e^\rho=0,\quad\sigma_{z,j}^2=\exp{(\bw_{\sigma_z^2,j\cdot})}=1,\quad q(z_j|\bx)=\mathcal{N}(0,1).
\end{equation}
The gradient of $\mu_{z,j}$ and $\sigma_{z,j}$ from $\mathcal{L}_{kl}(\bx;\phi)$ becomes
\begin{equation}
\frac{\partial\mathcal{L}_{kl}(\bx;\phi)}{\partial\mu_{z,j}}=\mu_{z,j}=0,\quad\frac{\partial\mathcal{L}_{kl}(\bx;\phi)}{\partial\sigma_{z,j}}=1-\sigma_{z,j}^{-1}=0.
\end{equation}
So the gradient of $\bw_{\mu_z,j\cdot}^\rho$ and $\bw_{\sigma_z^2,j\cdot}^\rho$ from $\mathcal{L}_{kl}$ is
\begin{equation}
\label{eqn:klloss_w_mu}
\frac{\partial\mathcal{L}_{kl}(\bx;\phi)}{\partial\bw_{\mu_z,j\cdot}^\rho}=\frac{\partial\mathcal{L}_{kl}(\bx;\phi)}{\partial\mu_{z,j}}{\bh_e^\rho}^\top=0,
\end{equation}
\begin{equation}
\label{eqn:klloss_w_sigma}
\frac{\partial\mathcal{L}_{kl}(\bx;\phi)}{\partial\bw_{\sigma_z^2,j\cdot}^\rho}=\frac{\partial\mathcal{L}_{kl}(\bx;\phi)}{2\sigma_{z,j}\cdot\partial\sigma_{z,j}}{\bh_e^\rho}^\top=0.
\end{equation}

\noindent Now we consider the gradient from $\mathcal{L}_{gen}(\bx;\theta,\phi)$. We have
\begin{equation}
	\frac{-\partial\log p_\theta(\bx|\bz)}{\partial z_j} = \frac{-\partial\log p_\theta(\bx|\bz)}{\partial\bh_d^1} \frac{\partial\bh_d^1}{\partial z_j}. \label{eqn:dz_from_gen}
\end{equation}
Since
\begin{equation}
	\bh_d^1 = \text{act} \left( \sum_{j=1}^\kappa \bw_{\mu_x,\cdot j}^1 z_j \right),
\end{equation}
where $\text{act}(\cdot)$ is the activation function, we can obtain
\begin{equation}
	\frac{\partial \bh_d^1}{\partial z_j} = \text{act}^\prime \left( \sum_{j=1}^\kappa \bw_{\mu_x,\cdot j}^1 z_j \right) \bw_{\mu_x,\cdot j}^1 = 0.
\end{equation}
Plugging this back into (\ref{eqn:dz_from_gen}) gives
\begin{equation}
	\frac{-\partial\log p_\theta(\bx|\bz)}{\partial z_j} = 0.
\end{equation}

According to the chain rule, we have
\begin{equation}
\label{eqn:exploss_w_mu}
\frac{\partial\mathcal{L}_{gen}(\bx;\theta,\phi)}{\partial\bw_{\mu_z,j\cdot}^\rho} = \mathbb{E}_{\bz\sim q_\phi(\bz|\bx)} \left[ \frac{-\partial\log p_\theta(\bx|\bz)}{\partial z_j}\frac{\partial z_j}{\partial\bw_{\mu_z,j\cdot}^\rho} \right] =0,
\end{equation}
\begin{equation}
\label{eqn:exploss_w_sigma}
\frac{\partial\mathcal{L}_{gen}(\bx;\theta,\phi)}{\partial\bw_{\sigma_z^2,j\cdot}^\rho} = \mathbb{E}_{\bz\sim q_\phi(\bz|\bx)} \left[ \frac{-\partial\log p_\theta(\bx|\bz)}{\partial z_j}\frac{\partial z_j}{\partial\bw_{\sigma_z^2,j\cdot}^\rho} \right] =0.
\end{equation}
After combining these two equations with (\ref{eqn:klloss_w_mu}) and (\ref{eqn:klloss_w_sigma}) and then integrating over $\bx$, we have
\begin{equation}
\frac{\partial\mathcal{L}(\theta,\phi)}{\partial\bw_{\mu_z,j\cdot}^\rho}=0,
\end{equation}
\begin{equation}
\frac{\partial\mathcal{L}(\theta,\phi)}{\partial\bw_{\sigma_z^2,j\cdot}^\rho}=0.
\end{equation}

Then we consider the gradient with respect to $\bw_{\mu_x,\cdot j}^1$. Since $\bw_{\mu_x,\cdot j}$ is part of $\theta$, it only receives gradient from $\mathcal{L}_{gen}(\bx;\theta,\phi)$. So we do not need to consider the KL loss. If $\bw_{\mu_x,\cdot j}^1=0$, $\bh_d^1=\sum_{j=1}^\kappa \bw_{\mu_x,\cdot j}^1z_j$ is not related to $\bz_j$. So $p_\theta(\bx|\bz)=p_\theta(\bx|\bz_{\neg j})$, where $\bz_{\neg j}$ represents $\bz$ without the $j$-th dimension. The gradient of $\bw_{\mu_x,\cdot j}^1$ is
\begin{align}
\frac{\partial\mathcal{L}_{gen}(\bx;\theta,\phi)}{\partial\bw_{\mu_x,\cdot j}^1}&=\mathbb{E}_{\bz\sim q(\bz|\bx)}\left[ \frac{-\partial\log p_\theta(\bx|\bz)}{\partial\bw_{\mu_x,\cdot j}^1} \right] = \mathbb{E}_{\bz\sim q(\bz|\bx)}\left[ \frac{-\partial\log p_\theta(\bx|\bz)}{\partial\bh_d^1} z_j \right]\\ \nonumber
&=\mathbb{E}_{\bz\neg j\sim q(\bz\neg j|\bx)} \left[ \mathbb{E}_{z_j\sim\mathcal{N}(0,1)} \left[ \frac{-\partial\log p_\theta(\bx|\bz_{\neg j})}{\partial\bh_d^1} z_j \right] \right]\\ \nonumber
&=\mathbb{E}_{\bz\neg i\sim q(\bz\neg i|\bx)}\left[\frac{-\partial \log p_\theta(\bx|\bz_{\neg j})}{\partial\bh_d^1} \mathbb{E}_{z_j\sim\mathcal{N}(0,1)}[\bz_j]\right]=0.
\end{align}
The integration over $\bx$ should also be $0$. So we obtain
\begin{equation}
	\frac{\partial\mathcal{L}(\theta;\phi)}{\partial \bw_{\mu_x,\cdot j}^1} = 0.
\end{equation}
\myendofproof 

\end{appendix}

\bibliography{refs,wipf_refs_nips2015_vers}
\bibliographystyle{iclr2020_conference}


\end{document}